\documentclass{article} %
\usepackage{iclr2022_conference,times}

\usepackage{amsmath,amsfonts,bm}

\def\eqref#1{equation~\ref{#1}}

\def\1{\bm{1}}

\DeclareMathAlphabet{\mathsfit}{\encodingdefault}{\sfdefault}{m}{sl}
\SetMathAlphabet{\mathsfit}{bold}{\encodingdefault}{\sfdefault}{bx}{n}

\usepackage{hyperref}
\usepackage{url}
\usepackage{tcolorbox}
\usepackage{hyperref}
\usepackage{wrapfig}
\usepackage{subcaption}
\usepackage{pifont}

\usepackage{multirow}
\usepackage{verbatim}
\usepackage{caption} 

\usepackage{enumitem}
\usepackage{amsthm}
\usepackage{amsmath}
\usepackage{amssymb}
\usepackage{etoolbox}
\usepackage{booktabs}
\usepackage{float}
\usepackage{makecell}

\newcommand{\parhead}[1]{\medskip \noindent {\bfseries\boldmath\ignorespaces #1.}\hskip 0.9em plus 0.3em minus 0.3em}

\definecolor{myred1}{RGB}{89, 10, 10}
\definecolor{myred2}{RGB}{145, 10, 10}
\definecolor{myred3}{RGB}{210, 10, 10}

\title{LoRA: Low-Rank Adaptation of Large Language Models}

\author{%
  Edward Hu\thanks{Equal contribution.} 
  \qquad Yelong Shen$^*$ 
  \qquad Phillip Wallis
  \qquad \textbf{Zeyuan Allen-Zhu} \\
  \textbf{Yuanzhi Li}
  \qquad \textbf{Shean Wang}
  \qquad \textbf{Lu Wang}
  \qquad \textbf{Weizhu Chen}\\
  Microsoft Corporation \\
  \texttt{\{edwardhu, yeshe, phwallis, zeyuana,} \\
  \texttt{yuanzhil, swang, luw, wzchen\}@microsoft.com} \\
  \texttt{yuanzhil@andrew.cmu.edu} \\
  (Version 2)}

\iclrfinalcopy
\begin{document}

\maketitle

\begin{abstract}
An important paradigm of natural language processing consists of large-scale pre-training on general domain data and adaptation to particular tasks or domains.
As we pre-train larger models, full fine-tuning, which retrains all model parameters, becomes less feasible.
Using GPT-3 175B as an example -- deploying independent instances of fine-tuned models, each with 175B parameters, is prohibitively expensive.
We propose \textbf{Lo}w-\textbf{R}ank \textbf{A}daptation, or LoRA, which freezes the pre-trained model weights and injects trainable rank decomposition matrices into each layer of the Transformer architecture, greatly reducing the number of trainable parameters for downstream tasks. 
Compared to GPT-3 175B fine-tuned with Adam, LoRA can reduce the number of trainable parameters by 10,000 times and the GPU memory requirement by 3 times.
LoRA performs on-par or better than fine-tuning in model quality on RoBERTa, DeBERTa, GPT-2, and GPT-3, despite having fewer trainable parameters, a higher training throughput, and, unlike adapters, \textit{no additional inference latency}.
We also provide an empirical investigation into rank-deficiency in language model adaptation, which sheds light on the efficacy of LoRA.
We release a package that facilitates the integration of LoRA with PyTorch models and provide our implementations and model checkpoints for RoBERTa, DeBERTa, and GPT-2 at~\url{https://github.com/microsoft/LoRA}.
\end{abstract}

\footnotetext{Compared to V1, this draft includes better baselines, experiments on GLUE, and more on adapter latency.}
\section{Introduction}

\begin{wrapfigure}{r}{0.29\textwidth}
  \centering
  \includegraphics[width=0.29\textwidth]{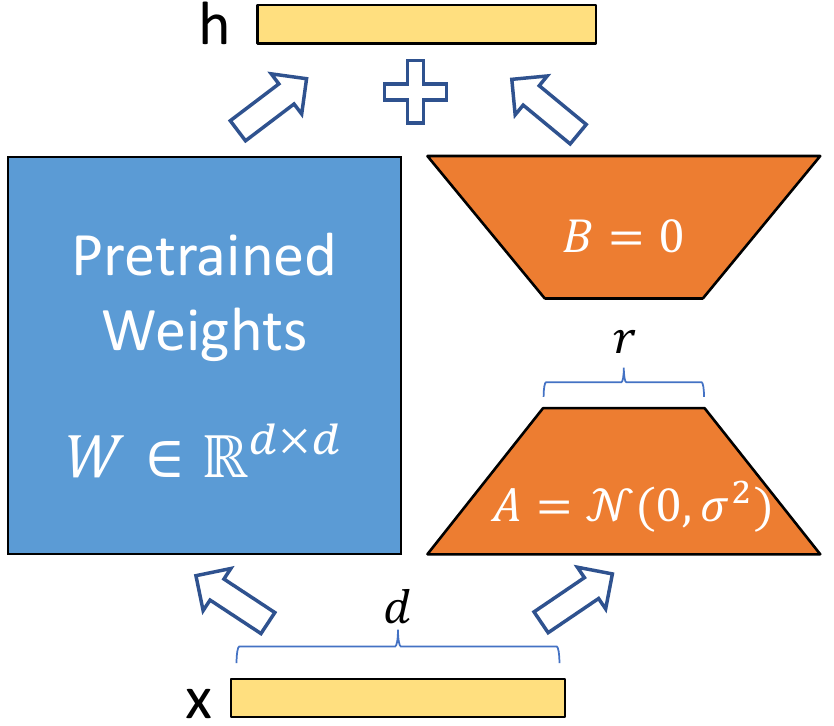}
  \caption{Our reparametrization. We only train $A$ and $B$.}
  \label{fig:reparam}
\end{wrapfigure}

Many applications in natural language processing rely on adapting \emph{one} large-scale, pre-trained language model to \emph{multiple} downstream applications.
Such adaptation is usually done via \emph{fine-tuning}, which updates all the parameters of the pre-trained model.
The major downside of fine-tuning is that the new model contains as many parameters as in the original model.
As larger models are trained every few months, this changes from a mere ``inconvenience'' for GPT-2~\citep{radford_language_nodate} or RoBERTa large~\citep{liu2019roberta} to a critical deployment challenge for GPT-3~\citep{brown_language_2020} with 175 billion trainable parameters.%
\footnote{While GPT-3 175B achieves non-trivial performance with few-shot learning, fine-tuning boosts its performance significantly as shown in~\autoref{app:fewshot_vs_finetune}.}

Many sought to mitigate this by adapting only some parameters or learning external modules for new tasks.
This way, we only need to store and load a small number of task-specific parameters in addition to the pre-trained model for each task, greatly boosting the operational efficiency when deployed.
However, existing techniques often introduce inference latency~\citep{houlsby_parameter-efficient_2019, rebuffi_learning_2017} by extending model depth or reduce the model's usable sequence length~\citep{li_prefix-tuning_2021, lester_power_2021, hambardzumyan_warp_2020, liu_gpt_2021} (\autoref{sec:existing_solutions_no_good}).
More importantly, these method often fail to match the fine-tuning baselines, posing a trade-off between efficiency and model quality.

We take inspiration from~\cite{li_measuring_2018, aghajanyan_intrinsic_2020} which show that the learned over-parametrized models in fact reside on a low intrinsic dimension.
We hypothesize that the change in weights during model adaptation also has a low ``intrinsic rank'', leading to our proposed \textbf{Lo}w-\textbf{R}ank \textbf{A}daptation (LoRA) approach.
LoRA allows us to train some dense layers in a neural network indirectly by optimizing rank decomposition matrices of the dense layers' change during adaptation instead, while keeping the pre-trained weights frozen, as shown in \autoref{fig:reparam}.
Using GPT-3 175B as an example, we show that a very low rank (i.e., \textit{r} in \autoref{fig:reparam} can be one or two) suffices even when the full rank (i.e., \textit{d}) is as high as 12,288, making LoRA both storage- and compute-efficient.

LoRA possesses several key advantages.
\begin{itemize}[topsep=6pt,itemsep=3pt,partopsep=4pt, parsep=4pt]
\item
A pre-trained model can be shared and used to build many small LoRA modules for different tasks.
We can freeze the shared model and efficiently switch tasks by replacing the matrices $A$ and $B$ in \autoref{fig:reparam}, reducing the storage requirement and task-switching overhead significantly.

\item
LoRA makes training more efficient and lowers the hardware barrier to entry by up to 3 times when using adaptive optimizers since we do not need to calculate the gradients or maintain the optimizer states for most parameters.
Instead, we only optimize the injected, much smaller low-rank matrices.%

\item
Our simple linear design allows us to merge the trainable matrices with the frozen weights when deployed, \textit{introducing no inference latency} compared to a fully fine-tuned model, by construction.

\item
LoRA is orthogonal to many prior methods and can be combined with many of them, such as prefix-tuning. We provide an example in \autoref{app:lora_plus}.

\end{itemize}

\paragraph{Terminologies and Conventions}
We make frequent references to the Transformer architecture and use the conventional terminologies for its dimensions.
We call the input and output dimension size of a Transformer layer $d_{model}$.
We use $W_q$, $W_k$, $W_v$, and $W_o$ to refer to the query/key/value/output projection matrices in the self-attention module.
$W$ or $W_0$ refers to a pre-trained weight matrix and $\Delta W$ its accumulated gradient update during adaptation.
We use $r$ to denote the rank of a LoRA module.
We follow the conventions set out by~\citep{vaswani2017attention, brown_language_2020} and use Adam~\citep{loshchilov2019decoupled, kingma2017adam} for model optimization and use a Transformer MLP feedforward dimension $d_{ffn} = 4 \times d_{model}$.

\section{Problem Statement}
\label{sec:problem_statement}
While our proposal is agnostic to training objective, we focus on language modeling as our motivating use case.
Below is a brief description of the language modeling problem and, in particular, the maximization of conditional probabilities given a task-specific prompt.

\newcommand{\cZ}{\mathcal{Z}}
Suppose we are given a pre-trained autoregressive language model $P_\Phi(y|x)$ parametrized by $\Phi$.
For instance, $P_\Phi(y|x)$ can be a generic multi-task learner such as GPT~\citep{radford_language_nodate,brown_language_2020} based on the Transformer architecture~\citep{vaswani2017attention}.
Consider adapting this pre-trained model to downstream conditional text generation tasks, such as summarization, machine reading comprehension (MRC), and natural language to SQL (NL2SQL).
Each downstream task is represented by a training dataset of context-target pairs: $\cZ = \{(x_i, y_i)\}_{i=1,..,N}$, where both $x_i$ and $y_i$ are sequences of tokens.
For example, in NL2SQL, $x_i$ is a natural language query and $y_i$ its corresponding SQL command; for summarization, $x_i$ is the content of an article and $y_i$ its summary.

During full fine-tuning, the model is initialized to pre-trained weights $\Phi_0$ and updated to $\Phi_0+\Delta \Phi$ by repeatedly following the gradient to maximize the conditional language modeling objective:
\begin{align}
    \max_{\Phi} \sum_{(x,y)\in\cZ} \sum_{t=1}^{|y|}  \text{log} \left(  P_{\Phi}(y_{t} | x, y_{<t}) \right)
\label{eq:ft_obj}
\end{align}
One of the main drawbacks for full fine-tuning is that for \emph{each} downstream task, we learn a \emph{different} set of parameters $\Delta\Phi$ whose dimension $|\Delta\Phi|$ equals $|\Phi_0|$. 
Thus, if the pre-trained model is large (such as GPT-3 with $|\Phi_0|\approx175 \text{~Billion}$), storing and deploying many independent instances of fine-tuned models can be challenging, if at all feasible. 

In this paper, we adopt a more parameter-efficient approach, where the task-specific parameter increment $\Delta\Phi = \Delta\Phi(\Theta)$ is further encoded by a much smaller-sized set of parameters $\Theta$ with $|\Theta| \ll |\Phi_0|$.
The task of finding $\Delta\Phi$ thus becomes optimizing over $\Theta$:
\begin{align}
    \max_{\Theta} \sum_{(x,y)\in\cZ}  \sum_{t=1}^{|y|}  \log\left({p_{\Phi_0+\Delta\Phi(\Theta)}(y_{t} | x, y_{<t}})\right)
\label{eq:ft_add}
\end{align}
In the subsequent sections, we propose to use a low-rank representation to encode $\Delta\Phi$ that is both compute- and memory-efficient.
When the pre-trained model is GPT-3 175B, the number of trainable parameters $|\Theta|$ can be as small as $0.01\%$ of $|\Phi_0|$. %

\section{Aren't Existing Solutions Good Enough?}
\label{sec:existing_solutions_no_good}

The problem we set out to tackle is by no means new.
Since the inception of transfer learning, dozens of works have sought to make model adaptation more parameter- and compute-efficient.
See~\autoref{sec:related_works} for a survey of some of the well-known works.
Using language modeling as an example, there are two prominent strategies when it comes to efficient adaptations: adding adapter layers~\citep{houlsby_parameter-efficient_2019, rebuffi_learning_2017, pfeiffer2021adapterfusion, ruckle2020adapterdrop} or optimizing some forms of the input layer activations~\citep{li_prefix-tuning_2021, lester_power_2021, hambardzumyan_warp_2020, liu_gpt_2021}.
However, both strategies have their limitations, especially in a large-scale and latency-sensitive production scenario.

\paragraph{Adapter Layers Introduce Inference Latency}
There are many variants of adapters.
We focus on the original design by~\citet{houlsby_parameter-efficient_2019} which has two adapter layers per Transformer block and a more recent one by ~\citet{lin-etal-2020-exploring} which has only one per block but with an additional LayerNorm~\citep{ba2016layer}.
While one can reduce the overall latency by pruning layers or exploiting multi-task settings~\citep{ruckle2020adapterdrop,pfeiffer2021adapterfusion}, there is no direct ways to bypass the extra compute in adapter layers.
This seems like a non-issue since adapter layers are designed to have few parameters (sometimes $<$1\% of the original model) by having a small bottleneck dimension, which limits the FLOPs they can add.
However, large neural networks rely on hardware parallelism to keep the latency low, and adapter layers have to be processed sequentially.
This makes a difference in the online inference setting where the batch size is typically as small as one.
In a generic scenario without model parallelism, such as running inference on GPT-2~\citep{radford_language_nodate} medium on a single GPU, we see a noticeable increase in latency when using adapters, even with a very small bottleneck dimension (\autoref{tab:adapter_slow}).

\begin{table}[h]
  \centering
  \begin{tabular}{c|ccc}
  \hline
  \toprule
  
  Batch Size              &  32                 & 16        & 1    \\
  Sequence Length         &  512                & 256       & 128  \\
  $|\Theta|$              &  0.5M         & 11M & 11M     \\
  \midrule
  Fine-Tune/LoRA               & 1449.4$\pm$0.8           & 338.0$\pm$0.6  & 19.8$\pm$2.7 \\
  \midrule
  $\text{Adapter}^{\text{L}}$   & 1482.0$\pm$1.0 (\textcolor{myred1}{+2.2\%})  & 354.8$\pm$0.5 (\textcolor{myred2}{+5.0\%})  & 23.9$\pm$2.1 (\textcolor{myred3}{+20.7\%}) \\
  $\text{Adapter}^{\text{H}}$   & 1492.2$\pm$1.0 (\textcolor{myred1}{+3.0\%})  & 366.3$\pm$0.5 (\textcolor{myred2}{+8.4\%})  & 25.8$\pm$2.2 (\textcolor{myred3}{+30.3\%}) \\
  \bottomrule
  \end{tabular}
  \caption{Infernece latency of a single forward pass in GPT-2 medium measured in milliseconds, averaged over 100 trials. We use an NVIDIA Quadro RTX8000. ``$|\Theta|$" denotes the number of trainable parameters in adapter layers. $\text{Adapter}^{\text{L}}$ and $\text{Adapter}^{\text{H}}$ are two variants of adapter tuning, which we describe in~\autoref{sec:expt_baselines}. The inference latency introduced by adapter layers can be significant in an online, short-sequence-length scenario. See the full study in~\autoref{app:adapter_latency}.}
  \label{tab:adapter_slow}
\end{table}

This problem gets worse when we need to shard the model as done in~\citet{shoeybi2020megatronlm, lepikhin2020gshard}, because the additional depth requires more synchronous GPU operations such as \texttt{AllReduce} and \texttt{Broadcast}, unless we store the adapter parameters redundantly many times.

\paragraph{Directly Optimizing the Prompt is Hard}
The other direction, as exemplified by prefix tuning~\citep{li_prefix-tuning_2021}, faces a different challenge.
We observe that prefix tuning is difficult to optimize and that its performance changes non-monotonically in trainable parameters, confirming similar observations in the original paper.
More fundamentally, reserving a part of the sequence length for adaptation necessarily reduces the sequence length available to process a downstream task, which we suspect makes tuning the prompt less performant compared to other methods.
We defer the study on task performance to~\autoref{sec:empirical}.

\section{Our Method}

We describe the simple design of LoRA and its practical benefits.
The principles outlined here apply to any dense layers in deep learning models, though we only focus on certain weights in Transformer language models in our experiments as the motivating use case.

\subsection{Low-Rank-Parametrized Update Matrices}
A neural network contains many dense layers which perform matrix multiplication.
The weight matrices in these layers typically have full-rank.
When adapting to a specific task, \citet{aghajanyan_intrinsic_2020} shows that the pre-trained language models have a low ``instrisic dimension'' and can still learn efficiently despite a random projection to a smaller subspace.
Inspired by this, we hypothesize the updates to the weights also have a low ``intrinsic rank" during adaptation.
For a pre-trained weight matrix $W_0\in \mathbb{R}^{d\times k}$, we constrain its update by representing the latter with a low-rank decomposition $W_0+\Delta W=W_0+BA$, where $B\in \mathbb{R}^{d\times r}, A\in \mathbb{R}^{r\times k}$, and the rank $r \ll \min(d,k)$.
During training, $W_0$ is frozen and does not receive gradient updates, while $A$ and $B$ contain trainable parameters.
Note both $W_0$ and $\Delta W=BA$ are multiplied with the same input, and their respective output vectors are summed coordinate-wise.
For $h = W_0x$, our modified forward pass yields:
\begin{equation}
h = W_0 x + \Delta W x = W_0 x + BA x
\label{eq:lora}
\end{equation}

We illustrate our reparametrization in \autoref{fig:reparam}.
We use a random Gaussian initialization for $A$ and zero for $B$, so $\Delta W=BA$ is zero at the beginning of training.
We then scale $\Delta W x$ by $\frac{\alpha}{r}$, where $\alpha$ is a constant in $r$.
When optimizing with Adam, tuning $\alpha$ is roughly the same as tuning the learning rate if we scale the initialization appropriately.
As a result, we simply set $\alpha$ to the first $r$ we try and do not tune it.
This scaling helps to reduce the need to retune hyperparameters when we vary $r$~\citep{yang_feature_2021}.

\parhead{A Generalization of Full Fine-tuning}
A more general form of fine-tuning allows the training of a subset of the pre-trained parameters.
LoRA takes a step further and does not require the accumulated gradient update to weight matrices to have full-rank during adaptation.
This means that when applying LoRA to all weight matrices and training all biases\footnote{They represent a negligible number of parameters compared to weights.}, we roughly recover the expressiveness of full fine-tuning by setting the LoRA rank $r$ to the rank of the pre-trained weight matrices.
In other words, as we increase the number of trainable parameters~\footnote{An inevitability when adapting to hard tasks.}, training LoRA roughly converges to training the original model, while adapter-based methods converges to an MLP and prefix-based methods to a model that cannot take long input sequences.

\parhead{No Additional Inference Latency}
When deployed in production, we can explicitly compute and store $W = W_0 + BA$ and perform inference as usual.
Note that both $W_0$ and $BA$ are in $\mathbb{R}^{d\times k}$.
When we need to switch to another downstream task, we can recover $W_0$ by subtracting $BA$ and then adding a different $B'A'$, a quick operation with very little memory overhead.
Critically, this guarantees that we do not introduce any additional latency during inference compared to a fine-tuned model by construction.

\subsection{Applying LoRA to Transformer}
\label{sec:apply_lora_to_tr}

In principle, we can apply LoRA to any subset of weight matrices in a neural network to reduce the number of trainable parameters.
In the Transformer architecture, there are four weight matrices in the self-attention module ($W_q, W_k, W_v, W_o$) and two in the MLP module.
We treat $W_q$ (or $W_k$, $W_v$) as a single matrix of dimension $d_{model} \times d_{model}$, even though the output dimension is usually sliced into attention heads.
We limit our study to \textbf{only adapting the attention weights} for downstream tasks and freeze the MLP modules (so they are not trained in downstream tasks) both for simplicity and parameter-efficiency.%
We further study the effect on adapting different types of attention weight matrices in a Transformer in \autoref{sec:weight_types}.
We leave the empirical investigation of adapting the MLP layers, LayerNorm layers, and biases to a future work.

\parhead{Practical Benefits and Limitations}
The most significant benefit comes from the reduction in memory and storage usage.
For a large Transformer trained with Adam, we reduce that VRAM usage by up to $2/3$ if $r\ll d_{model}$ as we do not need to store the optimizer states for the frozen parameters.
On GPT-3 175B, we reduce the VRAM consumption during training from 1.2TB to 350GB.
With $r=4$ and only the query and value projection matrices being adapted, the checkpoint size is reduced by roughly 10,000$\times$ (from 350GB to 35MB)\footnote{We still need the 350GB model during deployment; however, storing 100 adapted models only requires 350GB + 35MB * 100 $\approx$ 354GB as opposed to 100 * 350GB $\approx$ 35TB.}.
This allows us to train with significantly fewer GPUs and avoid I/O bottlenecks.
Another benefit is that we can switch between tasks while deployed at a much lower cost by only swapping the LoRA weights as opposed to all the parameters.
This allows for the creation of many customized models that can be swapped in and out on the fly on machines that store the pre-trained weights in VRAM.
We also observe a 25\% speedup during training on GPT-3 175B compared to full fine-tuning\footnote{For GPT-3 175B, the training throughput for full fine-tuning is 32.5 tokens/s per V100 GPU; with the same number of weight shards for model parallelism, the throughput is 43.1 tokens/s per V100 GPU for LoRA.} as we do not need to calculate the gradient for the vast majority of the parameters.

LoRA also has its limitations.
For example, it is not straightforward to batch inputs to different tasks with different $A$ and $B$ in a single forward pass, if one chooses to absorb $A$ and $B$ into $W$ to eliminate additional inference latency.
Though it is possible to not merge the weights and dynamically choose the LoRA modules to use for samples in a batch for scenarios where latency is not critical.

\section{Empirical Experiments}
\label{sec:empirical}
We evaluate the downstream task performance of LoRA on RoBERTa~\citep{liu2019roberta}, DeBERTa~\citep{he2021deberta}, and GPT-2~\citep{radford_language_nodate}, before scaling up to GPT-3 175B~\citep{brown_language_2020}.
Our experiments cover a wide range of tasks, from natural language understanding (NLU) to generation (NLG).
Specifically, we evaluate on the GLUE~\citep{wang2019glue} benchmark for RoBERTa and DeBERTa.
We follow the setup of~\citet{li_prefix-tuning_2021} on GPT-2 for a direct comparison and add WikiSQL~\citep{DBLP:journals/corr/abs-1709-00103} (NL to SQL queries) and SAMSum~\citep{DBLP:journals/corr/abs-1911-12237} (conversation summarization) for large-scale experiments on GPT-3.
See \autoref{app:datasets} for more details on the datasets we use.
We use NVIDIA Tesla V100 for all experiments.

\subsection{Baselines}
\label{sec:expt_baselines}
To compare with other baselines broadly, we replicate the setups used by prior work and reuse their reported numbers whenever possible.
This, however, means that some baselines might only appear in certain experiments.

\textbf{Fine-Tuning (FT)} is a common approach for adaptation.
During fine-tuning, the model is initialized to the pre-trained weights and biases, and all model parameters undergo gradient updates.%
A simple variant is to update only some layers while freezing others.
We include one such baseline reported in prior work~\citep{li_prefix-tuning_2021} on GPT-2, which adapts just the last two layers ($\textbf{FT}^{\textbf{Top2}}$).

\textbf{Bias-only or BitFit} is a baseline where we only train the bias vectors while freezing everything else.
Contemporarily, this baseline has also been studied by BitFit~\citep{zaken2021bitfit}.

\textbf{Prefix-embedding tuning (PreEmbed)} inserts special tokens among the input tokens.
These special tokens have trainable word embeddings and are generally not in the model's vocabulary.
Where to place such tokens can have an impact on performance.
We focus on ``prefixing'', which prepends such tokens to the prompt, and ``infixing'', which appends to the prompt; both are discussed in~\citet{li_prefix-tuning_2021}.
We use $l_{p}$ (resp. $l_{i}$) denote the number of prefix (resp. infix) tokens.
The number of trainable parameters is $|\Theta| = d_{model} \times (l_p + l_i)$. 

\textbf{Prefix-layer tuning (PreLayer)} is an extension to prefix-embedding tuning.
Instead of just learning the word embeddings (or equivalently, the activations after the embedding layer) for some special tokens, we learn the activations after every Transformer layer.
The activations computed from previous layers are simply replaced by trainable ones.
The resulting number of trainable parameters is $|\Theta| = L \times d_{model} \times (l_p + l_i)$, where $L$ is the number of Transformer layers.

\textbf{Adapter tuning} as proposed in~\citet{houlsby_parameter-efficient_2019} inserts adapter layers between the self-attention module (and the MLP module) and the subsequent residual connection.
There are two fully connected layers with biases in an adapter layer with a nonlinearity in between.
We call this original design $\textbf{Adapter}^{\textbf{H}}$.
Recently, \citet{lin-etal-2020-exploring} proposed a more efficient design with the adapter layer applied only after the MLP module and after a LayerNorm.
We call it $\textbf{Adapter}^{\textbf{L}}$.
This is very similar to another deign proposed in~\citet{pfeiffer2021adapterfusion}, which we call $\textbf{Adapter}^{\textbf{P}}$.
We also include another baseline call AdapterDrop~\citep{ruckle2020adapterdrop} which drops some adapter layers for greater efficiency ($\textbf{Adapter}^{\textbf{D}}$).
We cite numbers from prior works whenever possible to maximize the number of baselines we compare with; they are in rows with an asterisk (*) in the first column.
In all cases,  we have $|\Theta| = \hat{L}_{Adpt} \times (2 \times d_{model} \times r + r + d_{model}) + 2 \times \hat{L}_{LN} \times d_{model}$ where $\hat{L}_{Adpt}$ is the number of adapter layers and $\hat{L}_{LN}$ the number of trainable LayerNorms (e.g., in $\text{Adapter}^{\text{L}}$).

\textbf{LoRA} adds trainable pairs of rank decomposition matrices in parallel to existing weight matrices.
As mentioned in~\autoref{sec:apply_lora_to_tr}, we only apply LoRA to $W_{q}$ and $W_{v}$ in most experiments for simplicity.
The number of trainable parameters is determined by the rank $r$ and the shape of the original weights: $|\Theta| = 2 \times \hat{L}_{LoRA} \times d_{model} \times r$, where $\hat{L}_{LoRA}$ is the number of weight matrices we apply LoRA to.

\subsection{RoBERTa base/large}
\label{sec:bert_expt}

\begin{table}[t!]
  \centering
  \footnotesize
  \addtolength{\tabcolsep}{-4pt}
  \begin{tabular}{l|r|ccccccccc}
  \hline
  \toprule
  Model \& Method & \# Trainable & \multicolumn{9}{c}{} \\
         & Parameters & MNLI & SST-2 & MRPC & CoLA & QNLI & QQP & RTE & STS-B & Avg. \\
  \midrule
  $\text{RoB}_\text{base}$ (FT)* & 125.0M & \textbf{87.6} & 94.8 & 90.2 & \textbf{63.6} & 92.8 & \textbf{91.9} & 78.7 & 91.2 & 86.4 \\
  $\text{RoB}_\text{base}$ (BitFit)* & 0.1M & 84.7 & 93.7 & \textbf{92.7} & 62.0 & 91.8 & 84.0 & 81.5 & 90.8 & 85.2 \\
  $\text{RoB}_\text{base}$ ($\text{Adpt}^{\text{D}}$)* & 0.3M & 87.1\textsubscript{$\pm$.0} & 94.2\textsubscript{$\pm$.1} & 88.5\textsubscript{$\pm$1.1} & 60.8\textsubscript{$\pm$.4} & 93.1\textsubscript{$\pm$.1} & 90.2\textsubscript{$\pm$.0} & 71.5\textsubscript{$\pm$2.7} & 89.7\textsubscript{$\pm$.3} & 84.4 \\
  $\text{RoB}_\text{base}$ ($\text{Adpt}^{\text{D}}$)* & 0.9M & 87.3\textsubscript{$\pm$.1} & 94.7\textsubscript{$\pm$.3} & 88.4\textsubscript{$\pm$.1} & 62.6\textsubscript{$\pm$.9} & 93.0\textsubscript{$\pm$.2} & 90.6\textsubscript{$\pm$.0} & 75.9\textsubscript{$\pm$2.2} & 90.3\textsubscript{$\pm$.1} & 85.4 \\
  $\text{RoB}_\text{base}$ (LoRA) & 0.3M & 87.5\textsubscript{$\pm$.3} & \textbf{95.1\textsubscript{$\pm$.2}} & 89.7\textsubscript{$\pm$.7} & 63.4\textsubscript{$\pm$1.2} & \textbf{93.3\textsubscript{$\pm$.3}} & 90.8\textsubscript{$\pm$.1} & \textbf{86.6\textsubscript{$\pm$.7}} & \textbf{91.5\textsubscript{$\pm$.2}} & \textbf{87.2} \\
  \midrule
  $\text{RoB}_\text{large}$ (FT)* & 355.0M & 90.2 & \textbf{96.4} & \textbf{90.9} & 68.0 & 94.7 & \textbf{92.2} & 86.6 & 92.4 & 88.9 \\
  $\text{RoB}_\text{large}$ (LoRA) & 0.8M & \textbf{90.6}\textsubscript{$\pm$.2} & 96.2\textsubscript{$\pm$.5} & \textbf{90.9}\textsubscript{$\pm$1.2} & \textbf{68.2}\textsubscript{$\pm$1.9} & \textbf{94.9}\textsubscript{$\pm$.3} & 91.6\textsubscript{$\pm$.1} & \textbf{87.4}\textsubscript{$\pm$2.5} & \textbf{92.6}\textsubscript{$\pm$.2} & \textbf{89.0} \\
  \midrule
  $\text{RoB}_\text{large}$ ($\text{Adpt}^{\text{P}}$)$\dagger$ & 3.0M & 90.2\textsubscript{$\pm$.3} & 96.1\textsubscript{$\pm$.3} & 90.2\textsubscript{$\pm$.7} & \textbf{68.3}\textsubscript{$\pm$1.0} & \textbf{94.8}\textsubscript{$\pm$.2} & \textbf{91.9}\textsubscript{$\pm$.1} & 83.8\textsubscript{$\pm$2.9} & 92.1\textsubscript{$\pm$.7} & 88.4 \\
  $\text{RoB}_\text{large}$ ($\text{Adpt}^{\text{P}}$)$\dagger$ & 0.8M & \textbf{90.5}\textsubscript{$\pm$.3} & \textbf{96.6}\textsubscript{$\pm$.2} & 89.7\textsubscript{$\pm$1.2} & 67.8\textsubscript{$\pm$2.5} & \textbf{94.8}\textsubscript{$\pm$.3} & 91.7\textsubscript{$\pm$.2} & 80.1\textsubscript{$\pm$2.9} & 91.9\textsubscript{$\pm$.4} & 87.9 \\
  $\text{RoB}_\text{large}$ ($\text{Adpt}^{\text{H}}$)$\dagger$ & 6.0M & 89.9\textsubscript{$\pm$.5} & 96.2\textsubscript{$\pm$.3} & 88.7\textsubscript{$\pm$2.9} & 66.5\textsubscript{$\pm$4.4} & 94.7\textsubscript{$\pm$.2} & 92.1\textsubscript{$\pm$.1} & 83.4\textsubscript{$\pm$1.1} & 91.0\textsubscript{$\pm$1.7} & 87.8 \\
  $\text{RoB}_\text{large}$ ($\text{Adpt}^{\text{H}}$)$\dagger$ & 0.8M & 90.3\textsubscript{$\pm$.3} & 96.3\textsubscript{$\pm$.5} & 87.7\textsubscript{$\pm$1.7} & 66.3\textsubscript{$\pm$2.0} & 94.7\textsubscript{$\pm$.2} & 91.5\textsubscript{$\pm$.1} & 72.9\textsubscript{$\pm$2.9} & 91.5\textsubscript{$\pm$.5} & 86.4 \\
  $\text{RoB}_\text{large}$ (LoRA)$\dagger$ & 0.8M & \textbf{90.6}\textsubscript{$\pm$.2} & 96.2\textsubscript{$\pm$.5} & \textbf{90.2}\textsubscript{$\pm$1.0} & 68.2\textsubscript{$\pm$1.9} & \textbf{94.8}\textsubscript{$\pm$.3} & 91.6\textsubscript{$\pm$.2} & \textbf{85.2}\textsubscript{$\pm$1.1} & \textbf{92.3}\textsubscript{$\pm$.5} & \textbf{88.6} \\
  \midrule
  $\text{DeB}_\text{XXL}$ (FT)* & 1500.0M & 91.8 & \textbf{97.2} & 92.0 & 72.0 & \textbf{96.0} & 92.7 & 93.9 & 92.9 & 91.1 \\
  $\text{DeB}_\text{XXL}$ (LoRA) & 4.7M & \textbf{91.9}\textsubscript{$\pm$.2} & 96.9\textsubscript{$\pm$.2} & \textbf{92.6}\textsubscript{$\pm$.6} & \textbf{72.4}\textsubscript{$\pm$1.1} & \textbf{96.0}\textsubscript{$\pm$.1} & \textbf{92.9}\textsubscript{$\pm$.1} & \textbf{94.9}\textsubscript{$\pm$.4} & \textbf{93.0}\textsubscript{$\pm$.2} & \textbf{91.3} \\
  \bottomrule
  \end{tabular}
  \caption{$\text{RoBERTa}_{\text{base}}$, $\text{RoBERTa}_{\text{large}}$, and $\text{DeBERTa}_{\text{XXL}}$ with different adaptation methods on the GLUE benchmark. We report the overall (matched and mismatched) accuracy for MNLI, Matthew's correlation for CoLA, Pearson correlation for STS-B, and accuracy for other tasks. Higher is better for all metrics. * indicates numbers published in prior works. $\dagger$ indicates runs configured in a setup similar to~\cite{houlsby_parameter-efficient_2019} for a fair comparison.
  }
  \label{tab:NLU_results}
\end{table}
RoBERTa~\citep{liu2019roberta} optimized the pre-training recipe originally proposed in BERT~\citep{devlin2019bert} and boosted the latter's task performance without introducing many more trainable parameters.
While RoBERTa has been overtaken by much larger models on NLP leaderboards such as the GLUE benchmark~\citep{wang2019glue} in recent years, it remains a competitive and popular pre-trained model for its size among practitioners.
We take the pre-trained RoBERTa base (125M) and RoBERTa large (355M) from the HuggingFace Transformers library~\citep{wolf-etal-2020-transformers} and evaluate the performance of different efficient adaptation approaches on tasks from the GLUE benchmark.
We also replicate~\cite{houlsby_parameter-efficient_2019} and~\cite{pfeiffer2021adapterfusion} according to their setup.
To ensure a fair comparison, we make two crucial changes to how we evaluate LoRA when comparing with adapters.
First, we use the same batch size for all tasks and use a sequence length of 128 to match the adapter baselines.
Second, we initialize the model to the pre-trained model for MRPC, RTE, and STS-B, not a model already adapted to MNLI like the fine-tuning baseline.
Runs following this more restricted setup from~\cite{houlsby_parameter-efficient_2019} are labeled with $\dagger$.
The result is presented in~\autoref{tab:NLU_results} (Top Three Sections).
See~\autoref{app:hps_roberta} for details on the hyperparameters used.

\subsection{DeBERTa XXL}
\label{sec:deberta_expt}
DeBERTa~\citep{he2021deberta} is a more recent variant of BERT that is trained on a much larger scale and performs very competitively on benchmarks such as GLUE~\citep{wang2019glue} and SuperGLUE~\citep{wang2020superglue}.
We evaluate if LoRA can still match the performance of a fully fine-tuned DeBERTa XXL (1.5B) on GLUE.
The result is presented in~\autoref{tab:NLU_results} (Bottom Section).
See~\autoref{app:hps_deberta} for details on the hyperparameters used.

\subsection{GPT-2 medium/large}
\label{sec:gpt2_expt}
Having shown that LoRA can be a competitive alternative to full fine-tuning on NLU, we hope to answer if LoRA still prevails on NLG models, such as GPT-2 medium and large~\citep{radford_language_nodate}.
We keep our setup as close as possible to~\cite{li_prefix-tuning_2021} for a direct comparison.
Due to space constraint, we only present our result on E2E NLG Challenge (\autoref{tab:gpt2_ft_results}) in this section.
See~\autoref{app:gpt2_extra} for results on WebNLG~\citep{gardent2017webnlg} and DART~\citep{nan2020dart}.
We include a list of the hyperparameters used in~\autoref{app:hps_gpt2}.

\begin{table}[t]
\centering
\begin{tabular}{l|r|ccccc}
\hline
\toprule
Model \& Method & \# Trainable & \multicolumn{5}{c}{E2E NLG Challenge} \\
       & Parameters & BLEU & NIST & MET & ROUGE-L & CIDEr \\
\midrule
GPT-2 M (FT)* & 354.92M                         & 68.2 &	8.62 &	46.2 &	71.0 &	2.47  \\
GPT-2 M ($\text{Adapter}^{\text{L}}$)* & 0.37M  & 66.3 &	8.41 &	45.0 &	69.8 &	2.40  \\
GPT-2 M ($\text{Adapter}^{\text{L}}$)* & 11.09M & 68.9 &	8.71 &	46.1 &	71.3 &	2.47  \\
GPT-2 M ($\text{Adapter}^{\text{H}}$) & 11.09M & 67.3\textsubscript{$\pm$.6} & 8.50\textsubscript{$\pm$.07}	& 46.0\textsubscript{$\pm$.2} & 70.7\textsubscript{$\pm$.2}	& 2.44\textsubscript{$\pm$.01}        \\
GPT-2 M ($\text{FT}^{\text{Top2}}$)*   & 25.19M & 68.1 & 8.59 & 46.0  &  70.8 & 2.41  \\
GPT-2 M (PreLayer)* & 0.35M & 69.7 & 8.81 & 46.1 & 71.4 & 2.49  \\
GPT-2 M (LoRA) & 0.35M & \textbf{70.4\textsubscript{$\pm$.1}} & \textbf{8.85\textsubscript{$\pm$.02}} & \textbf{46.8\textsubscript{$\pm$.2}} & \textbf{71.8\textsubscript{$\pm$.1}} & \textbf{2.53\textsubscript{$\pm$.02}} \\
\midrule
GPT-2 L (FT)* & 774.03M & 68.5 & 8.78 & 46.0 & 69.9 & 2.45  \\
GPT-2 L ($\text{Adapter}^{\text{L}}$) & 0.88M  & 69.1\textsubscript{$\pm$.1} & 8.68\textsubscript{$\pm$.03} & 46.3\textsubscript{$\pm$.0} & 71.4\textsubscript{$\pm$.2} &	\textbf{2.49\textsubscript{$\pm$.0}}  \\
GPT-2 L ($\text{Adapter}^{\text{L}}$) & 23.00M & 68.9\textsubscript{$\pm$.3} & 8.70\textsubscript{$\pm$.04} & 46.1\textsubscript{$\pm$.1} & 71.3\textsubscript{$\pm$.2} &   2.45\textsubscript{$\pm$.02}  \\
GPT-2 L (PreLayer)* & 0.77M & 70.3 & 8.85 & 46.2 & 71.7 & 2.47  \\
GPT-2 L (LoRA) & 0.77M & \textbf{70.4\textsubscript{$\pm$.1}} & \textbf{8.89\textsubscript{$\pm$.02}} & \textbf{46.8\textsubscript{$\pm$.2}} & \textbf{72.0\textsubscript{$\pm$.2}} & 2.47\textsubscript{$\pm$.02} \\
\bottomrule
\end{tabular}
\caption{GPT-2 medium (M) and large (L) with different adaptation methods on the E2E NLG Challenge. For all metrics, higher is better. LoRA outperforms several baselines with comparable or fewer trainable parameters. Confidence intervals are shown for experiments we ran. * indicates numbers published in prior works.
}
\label{tab:gpt2_ft_results}
\end{table}

\begin{table}[h]
  \centering
  \begin{tabular}{l|r|ccc}
  \hline
  \toprule
  \multirow{2}{*}{Model\&Method}  & \# Trainable &  WikiSQL & MNLI-m & SAMSum  \\ %
  \cline{3-5}
  & Parameters & Acc. (\%) & Acc. (\%) & R1/R2/RL \\
  \midrule
  GPT-3 (FT)                         & 175,255.8M &  \textbf{73.8}  &  89.5 & 52.0/28.0/44.5 \\
  GPT-3 (BitFit)                  & 14.2M & 71.3 & 91.0 & 51.3/27.4/43.5 \\
  GPT-3 (PreEmbed)                   & 3.2M  & 63.1 & 88.6 & 48.3/24.2/40.5 \\
  GPT-3 (PreLayer)                   & 20.2M & 70.1 & 89.5 & 50.8/27.3/43.5 \\
  GPT-3 ($\text{Adapter}^{\text{H}}$)& 7.1M  & 71.9 & 89.8 & 53.0/28.9/44.8  \\
  GPT-3 ($\text{Adapter}^{\text{H}}$)& 40.1M & 73.2 & \textbf{91.5} & 53.2/29.0/45.1  \\
  \midrule
  GPT-3 (LoRA)                       & 4.7M & 73.4 & \textbf{91.7} & \textbf{53.8/29.8/45.9} \\ 
  GPT-3 (LoRA)                       & 37.7M & \textbf{74.0} & \textbf{91.6} & 53.4/29.2/45.1 \\ 
  \bottomrule
  \end{tabular}
  \caption{Performance of different adaptation methods on GPT-3 175B. We report the logical form validation accuracy on WikiSQL, validation accuracy on MultiNLI-matched, and Rouge-1/2/L on SAMSum. LoRA performs better than prior approaches, including full fine-tuning. The results on WikiSQL have a fluctuation around $\pm0.5\%$, MNLI-m around $\pm0.1\%$, and SAMSum around $\pm0.2$/$\pm0.2$/$\pm0.1$ for the three metrics.}
  \label{tab:gpt3_ft_results}
\end{table}

\subsection{Scaling up to GPT-3 175B}
\label{sec:gpt3_expts}
As a final stress test for LoRA, we scale up to GPT-3 with 175 billion parameters.
Due to the high training cost, we only report the typical standard deviation for a given task over random seeds, as opposed to providing one for every entry.
See~\autoref{app:hps_gpt3} for details on the hyperparameters used.

\begin{figure}[h]
\centering
\includegraphics[width=0.97\textwidth]{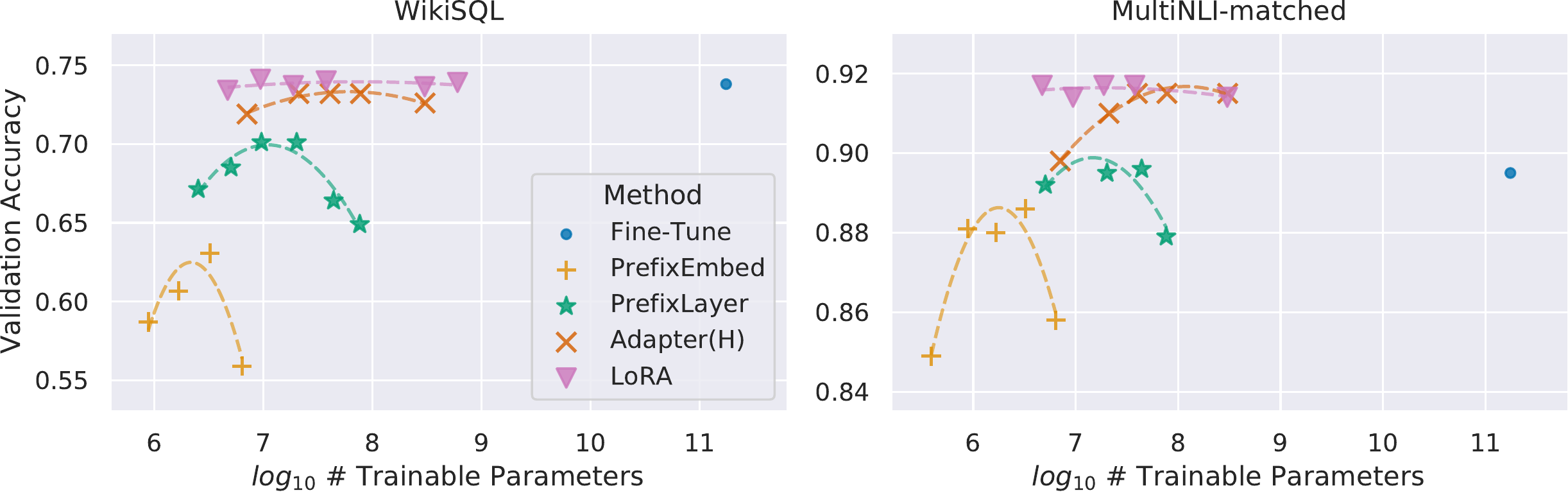}
\caption{GPT-3 175B validation accuracy vs. number of trainable parameters of several adaptation methods on WikiSQL and MNLI-matched. LoRA exhibits better scalability and task performance. See~\autoref{app:gpt3_extra} for more details on the plotted data points.}
\label{fig:eff}
\end{figure}

As shown in \autoref{tab:gpt3_ft_results}, LoRA matches or exceeds the fine-tuning baseline on all three datasets.
Note that not all methods benefit monotonically from having more trainable parameters, as shown in~\autoref{fig:eff}.
We observe a significant performance drop when we use more than 256 special tokens for prefix-embedding tuning or more than 32 special tokens for prefix-layer tuning.
This corroborates similar observations in~\cite{li_prefix-tuning_2021}.
While a thorough investigation into this phenomenon is out-of-scope for this work, we suspect that having more special tokens causes the input distribution to shift further away from the pre-training data distribution.
Separately, we investigate the performance of different adaptation approaches in the low-data regime in~\autoref{app:low_data}.

\section{Related Works}
\label{sec:related_works}

\parhead{Transformer Language Models}
Transformer~\citep{vaswani2017attention} is a sequence-to-sequence architecture that makes heavy use of self-attention.
\citet{radford_improving_nodate} applied it to autoregressive language modeling by using a stack of Transformer decoders.
Since then, Transformer-based language models have dominated NLP, achieving the state-of-the-art in many tasks.
A new paradigm emerged with BERT~\citep{devlin_bert_2019} and GPT-2~\citep{radford_language_nodate} -- both are large Transformer language models trained on a large amount of text -- where fine-tuning on task-specific data after pre-training on general domain data provides a significant performance gain compared to training on task-specific data directly.
Training larger Transformers generally results in better performance and remains an active research direction.
GPT-3~\citep{brown_language_2020} is the largest single Transformer language model trained to-date with 175B parameters.

\parhead{Prompt Engineering and Fine-Tuning}
While GPT-3 175B can adapt its behavior with just a few additional training examples, the result depends heavily on the input prompt~\citep{brown_language_2020}.
This necessitates an empirical art of composing and formatting the prompt to maximize a model's performance on a desired task, which is known as prompt engineering or prompt hacking.
Fine-tuning retrains a model pre-trained on general domains to a specific task~\cite{devlin_bert_2019,radford_improving_nodate}.
Variants of it include learning just a subset of the parameters~\cite{devlin_bert_2019,collobert_unified_2008}, yet practitioners often retrain all of them to maximize the downstream performance.
However, the enormity of GPT-3 175B makes it challenging to perform fine-tuning in the usual way due to the large checkpoint it produces and the high hardware barrier to entry since it has the same memory footprint as pre-training.

\parhead{Parameter-Efficient Adaptation}
Many have proposed inserting \textit{adapter} layers between existing layers in a neural network~\citep{houlsby_parameter-efficient_2019, rebuffi_learning_2017,lin-etal-2020-exploring}.
Our method uses a similar bottleneck structure to impose a low-rank constraint on the weight updates.
The key functional difference is that our learned weights can be merged with the main weights during inference, thus not introducing any latency, which is not the case for the adapter layers (\autoref{sec:existing_solutions_no_good}).
A comtenporary extension of adapter is \textsc{compacter}~\citep{mahabadi2021compacter}, which essentially parametrizes the adapter layers using Kronecker products with some predetermined weight sharing scheme.
Similarly, combining LoRA with other tensor product-based methods could potentially improve its parameter efficiency, which we leave to future work.
More recently, many proposed optimizing the input word embeddings in lieu of fine-tuning, akin to a continuous and differentiable generalization of prompt engineering~\citep{li_prefix-tuning_2021, lester_power_2021, hambardzumyan_warp_2020, liu_gpt_2021}.
We include comparisons with~\citet{li_prefix-tuning_2021} in our experiment section.
However, this line of works can only scale up by using more special tokens in the prompt, which take up available sequence length for task tokens when positional embeddings are learned.

\parhead{Low-Rank Structures in Deep Learning} Low-rank structure is very common in machine learning.
A lot of machine learning problems have certain intrinsic low-rank structure~\citep{li2016recovery,cai2010singular,li2018algorithmic,grasedyck2013literature}.
Moreover, it is known that for many deep learning tasks, especially those with a heavily over-parametrized neural network, the learned neural network will enjoy low-rank properties after training~\citep{oymak2019generalization}.
Some prior works even explicitly impose the low-rank constraint when training the original neural network~\citep{sainath2013low,povey2018semi,zhang2014extracting,jaderberg2014speeding,zhao2016low,khodak2021initialization,denil2014predicting}; however, to the best of our knowledge, none of these works considers low-rank update to a frozen model for \emph{adaptation to downstream tasks}.
In theory literature, it is known that neural networks outperform other classical learning methods, including the corresponding (finite-width) neural tangent kernels~\citep{als18dnn,li2018learning} when the underlying concept class has certain low-rank structure~\citep{ghorbani2020neural,AL2019-resnet,allen2020backward}. 
Another theoretical result in \citet{allen2020feature} suggests that low-rank adaptations can be useful for adversarial training.
In sum, we believe that our proposed low-rank adaptation update is well-motivated by the literature. 

\section{Understanding the Low-Rank Updates}
\label{sec:science}

Given the empirical advantage of LoRA, we hope to further explain the properties of the low-rank adaptation learned from downstream tasks.
Note that the low-rank structure not only lowers the hardware barrier to entry which allows us to run multiple experiments in parallel, but also gives better interpretability of how the update weights are correlated with the pre-trained weights.
We focus our study on GPT-3 175B, where we achieved the largest reduction of trainable parameters (up to 10,000$\times$) without adversely affecting task performances.

We perform a sequence of empirical studies to answer the following questions:
1) Given a parameter budget constraint, \emph{which subset of weight matrices} in a pre-trained Transformer should we adapt to maximize downstream performance?
2) Is the ``optimal'' adaptation matrix $\Delta W$ \emph{really rank-deficient}? If so, what is a good rank to use in practice?
3) What is the connection between $\Delta W$ and $W$? Does $\Delta W$ highly correlate with $W$? How large is $\Delta W$ comparing to $W$?

We believe that our answers to question (2) and (3) shed light on the fundamental principles of using pre-trained language models for downstream tasks, which is a critical topic in NLP.

\subsection{Which Weight Matrices in Transformer Should We Apply LoRA to?}
\label{sec:weight_types}
Given a limited parameter budget, which types of weights should we adapt with LoRA to obtain the best performance on downstream tasks?
As mentioned in~\autoref{sec:apply_lora_to_tr}, we only consider weight matrices in the self-attention module.
We set a parameter budget of 18M (roughly 35MB if stored in FP16) on GPT-3 175B, which corresponds to $r=8$ if we adapt one type of attention weights or $r=4$ if we adapt two types, for all 96 layers.
The result is presented in \autoref{tab:weight_type}.

\begin{table}[h]
  \centering
  \begin{tabular}{l|ccccccc}
  \hline
  \toprule
                & \multicolumn{7}{c}{\# of Trainable Parameters = 18M} \\
  \midrule
  Weight Type           & $W_q$  & $W_k$  & $W_v$  & $W_o$      & $W_q,W_k$     & $W_q,W_v$     & $W_q, W_k, W_v, W_o$      \\
  Rank $r$              & 8      &  8     &  8     &   8        &   4           &   4           &   2                       \\
  \midrule
  WikiSQL ($\pm0.5$\%)  & 70.4   & 70.0   & 73.0   & 73.2       & 71.4          & \textbf{73.7} & \textbf{73.7}             \\
  MultiNLI ($\pm0.1$\%) & 91.0   & 90.8   & 91.0   & 91.3       & 91.3          & 91.3          & \textbf{91.7}              \\
  \bottomrule
  \end{tabular}
  \caption{Validation accuracy on WikiSQL and MultiNLI after applying LoRA to different types of attention weights in GPT-3, given the same number of trainable parameters. Adapting both $W_q$ and $W_v$ gives the best performance overall. We find the standard deviation across random seeds to be consistent for a given dataset, which we report in the first column.}
  \label{tab:weight_type}
\end{table}

Note that putting all the parameters in $\Delta W_q$ or $\Delta W_k$ results in significantly lower performance, while adapting both $W_q$ and $W_v$ yields the best result.
This suggests that even a rank of four captures enough information in $\Delta W$ such that it is preferable to adapt more weight matrices than adapting a single type of weights with a larger rank.

\subsection{What is the Optimal Rank $r$ for LoRA?}
\label{sec:effect_of_r}

We turn our attention to the effect of rank $r$ on model performance.
We adapt $\{W_q, W_v\}$, $\{W_q, W_k, W_v, W_c\}$, and just $W_q$ for a comparison.

\begin{table}[h]
  \centering
  \begin{tabular}{r|c|ccccc}
  \hline
  \toprule
                                        & Weight Type           & $r=1$  & $r=2$  & $r=4$  & $r=8$  & $r=64$  \\
  \midrule
  \multirow{2}{*}{WikiSQL($\pm0.5$\%)}  & $W_{q}$               & 68.8   & 69.6   & 70.5   & 70.4   & 70.0    \\
                                        & $W_q, W_v$            & 73.4   & 73.3   & 73.7   & 73.8   & 73.5    \\
                                        & $W_q, W_k, W_v, W_o$  & 74.1   & 73.7   & 74.0   & 74.0   & 73.9    \\
  \midrule
  \multirow{3}{*}{MultiNLI ($\pm0.1$\%)}& $W_q$                 & 90.7   & 90.9   & 91.1   & 90.7   & 90.7    \\ 
                                        & $W_q, W_v$            & 91.3   & 91.4   & 91.3   & 91.6   & 91.4    \\ 
                                        & $W_q, W_k, W_v, W_o$  & 91.2   & 91.7   & 91.7   & 91.5   & 91.4    \\
  
  \bottomrule
  \end{tabular}
  \caption{Validation accuracy on WikiSQL and MultiNLI with different rank $r$. To our surprise, a rank as small as one suffices for adapting both $W_q$ and $W_v$ on these datasets while training $W_q$ alone needs a larger $r$. We conduct a similar experiment on GPT-2 in~\autoref{app:gpt2_effect_r}.}
  \label{tab:effect_r}
\end{table}

\autoref{tab:effect_r} shows that, surprisingly, LoRA already performs competitively with a very small $r$ (more so for $\{W_q, W_v\}$ than just $W_q$).
This suggests the update matrix $\Delta W$ could have a very small ``intrinsic rank".%
\footnote{However, we do not expect a small $r$ to work for every task or dataset.
Consider the following thought experiment: if the downstream task were in a different language than the one used for pre-training, retraining the entire model (similar to LoRA with $r=d_{model}$) could certainly outperform LoRA with a small $r$.}
To further support this finding, we check the overlap of the subspaces learned by different choices of $r$ and by different random seeds. We argue that increasing $r$ does not cover a more meaningful subspace, which suggests that a low-rank adaptation matrix is sufficient.

\parhead{Subspace similarity between different $r$}
Given $A_{r=8}$ and $A_{r=64}$ which are the learned adaptation matrices with rank $r = 8$ and $64$ using the \emph{same pre-trained model}, we perform singular value decomposition and obtain the right-singular unitary matrices $U_{A_{r=8}}$ and $U_{A_{r=64}}$.%
\footnote{Note that a similar analysis can be carried out with $B$ and the left-singular unitary matrices -- we stick with $A$ for our experiments.}
We hope to answer: how much of the subspace spanned by the top $i$ singular vectors in $U_{A_{r=8}}$ (for $1\leq i \leq 8$) is contained in the subspace spanned by top $j$ singular vectors of $U_{A_{r=64}}$  (for $1\leq j \leq 64$)?
We measure this quantity with a normalized subspace similarity based on the Grassmann distance (See~\autoref{app:grassmann_distance} for a more formal discussion)
\begin{equation}
  \phi(A_{r=8}, A_{r=64}, i, j) = \frac{||U_{A_{r=8}}^{i\top} U_{A_{r=64}}^j||_{F}^2}{\min(i, j)} \in [0,1]
  \label{eq:overlap}
\end{equation}

where $U_{A_{r=8}}^i$ represents the columns of $U_{A_{r=8}}$ corresponding to the top-$i$ singular vectors.

\begin{figure}[h]
  \centering
    \includegraphics[width=0.97\textwidth]{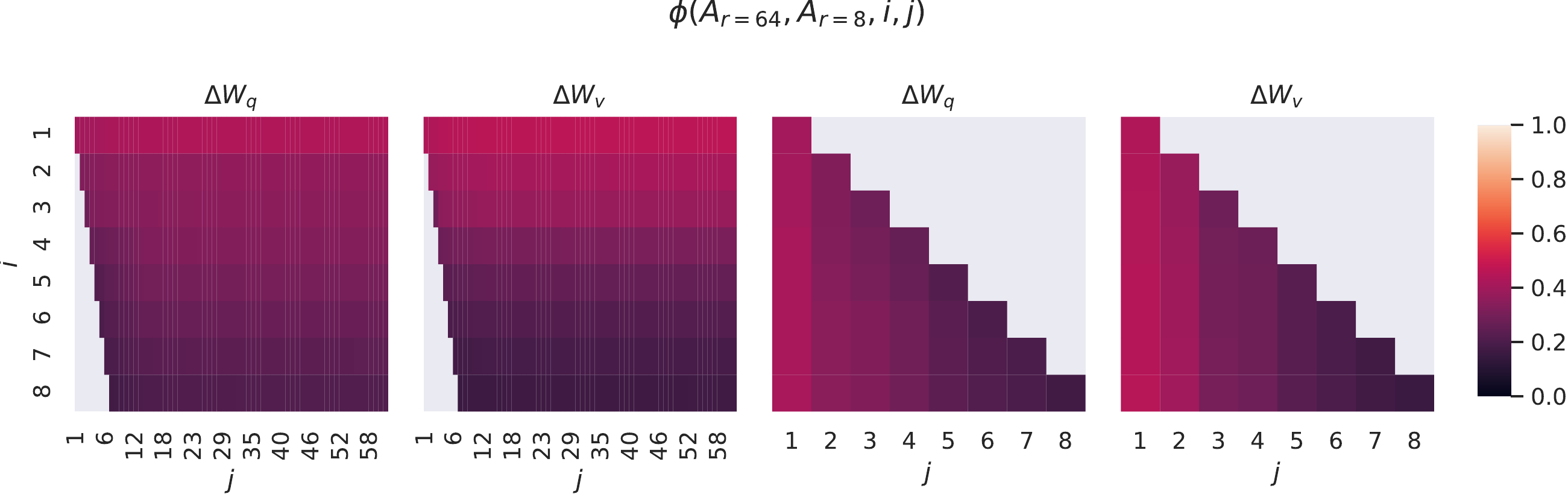}
    \caption{Subspace similarity between column vectors of $A_{r=8}$ and $A_{r=64}$ for both $\Delta W_q$ and $\Delta W_v$. The third and the fourth figures zoom in on the lower-left triangle in the first two figures. The top directions in $r=8$ are included in $r=64$, and vice versa.}
    \label{fig:qv8_qv64}
\end{figure}

$\phi(\cdot)$ has a range of $[0, 1]$, where $1$ represents a complete overlap of subspaces and $0$ a complete separation.
See \autoref{fig:qv8_qv64} for how $\phi$ changes as we vary $i$ and $j$.
We only look at the 48th layer (out of 96) due to space constraint, but the conclusion holds for other layers as well, as shown in~\autoref{app:corr_lora}.

We make an \emph{important observation} from \autoref{fig:qv8_qv64}.

\begin{center}
\parbox{0.85\linewidth}{
Directions corresponding to the top singular vector overlap significantly between $A_{r=8}$ and $A_{r=64}$, while others do not. Specifically, $\Delta W_v$ (resp. $\Delta W_q$) of $A_{r=8}$ and $\Delta W_v$ (resp. $\Delta W_q$) of $A_{r=64}$ share a subspace of dimension 1 with normalized similarity $>0.5$, providing an explanation of why $r=1$ performs quite well in our downstream tasks for GPT-3.}
\end{center}

Since both  $A_{r=8}$ and $A_{r=64}$ are learned using the same pre-trained model, \autoref{fig:qv8_qv64} indicates that the top singular-vector directions of $A_{r=8}$ and $A_{r=64}$ are the most useful, while other directions potentially contain mostly random noises accumulated during training.
Hence, the adaptation matrix can indeed have a very low rank.

\begin{figure}[h]
  \centering
    \includegraphics[width=0.92\textwidth]{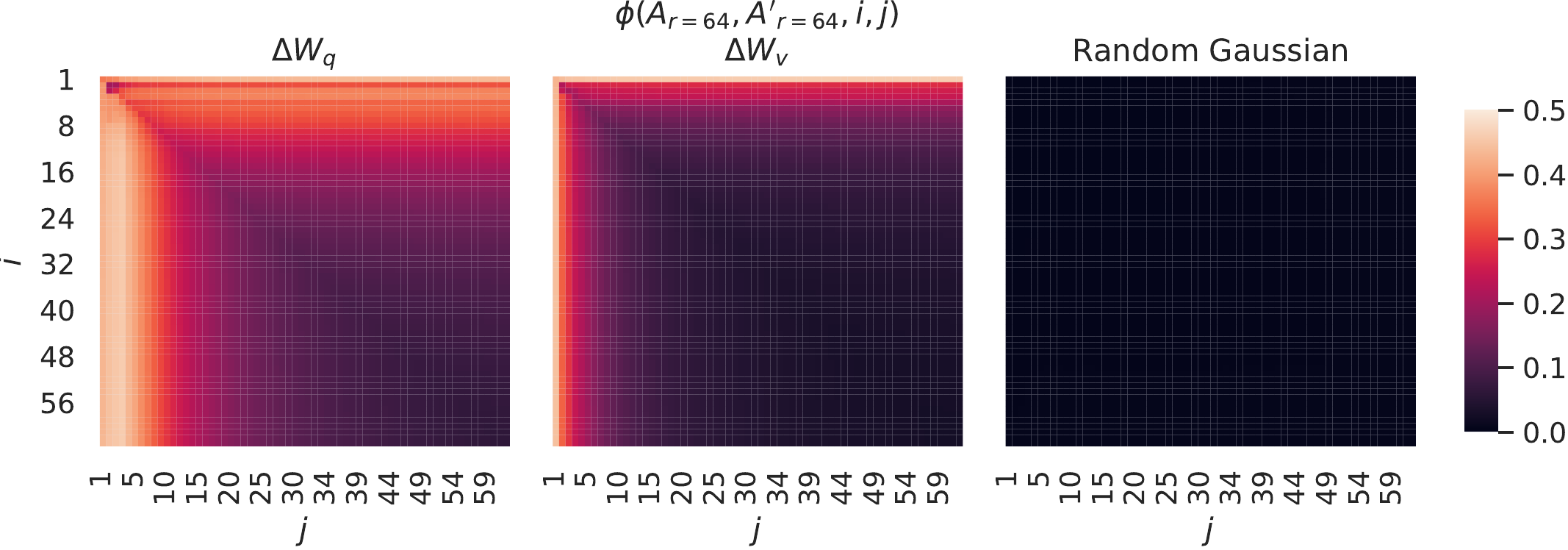}
    \caption{\textbf{Left and Middle:} Normalized subspace similarity between the column vectors of $A_{r=64}$ from two random seeds, for both $\Delta W_q$ and $\Delta W_v$ in the 48-th layer. \textbf{Right:} the same heat-map between the column vectors of two random Gaussian matrices. See~\autoref{app:corr_lora} for other layers.}
    \label{fig:UA_across_random_seeds}
\end{figure}

\parhead{Subspace similarity between different random seeds}
We further confirm this by plotting the normalized subspace similarity between two randomly seeded runs with $r=64$, shown in \autoref{fig:UA_across_random_seeds}.
$\Delta W_q$ appears to have a higher ``intrinsic rank'' than $\Delta W_v$, since more common singular value directions are learned by both runs for $\Delta W_q$, which is in line with our empirical observation in~\autoref{tab:effect_r}.
As a comparison, we also plot two random Gaussian matrices, which do not share any common singular value directions with each other.

\subsection{How Does the Adaptation Matrix $\Delta W$ Compare to $W$?}
\label{sec:compare_delta_w_to_w}
We further investigate the relationship between $\Delta W$ and $W$.
In particular, does $\Delta W$ highly correlate with $W$? (Or mathematically, is $\Delta W$ mostly contained in the top singular directions of $W$?) Also, how ``large'' is $\Delta W$ comparing to its corresponding directions in $W$?
This can shed light on the underlying mechanism for adapting pre-trained language models. 

To answer these questions, we project $W$ onto the $r$-dimensional subspace of $\Delta W$ by computing $U^\top W V^\top$, with $U$/$V$ being the left/right singular-vector matrix of $\Delta W$. Then, we compare the Frobenius norm between $\|U^\top W V^\top\|_F$ and $\|W\|_F$. 
As a comparison, we also compute $\|U^\top W V^\top\|_F$ by replacing $U,V$ with the top $r$ singular vectors of $W$ or a random matrix.

\begin{table}[h]
  \centering
  \begin{tabular}{c|ccc|ccc}
  \hline
  \toprule
                             & \multicolumn{3}{c|}{$r=4$}             & \multicolumn{3}{c}{$r=64$}           \\
                                 & $\Delta W_q$ & $W_q$ & Random  &  $\Delta W_q$ & $W_q$  &  Random \\
  \midrule
  $||U^\top W_qV^\top||_F = $        & 0.32   & 21.67     & 0.02     & 1.90  & 37.71                & 0.33    \\
  \midrule
  $||W_q||_F = 61.95$                & \multicolumn{3}{c|}{$||\Delta W_q||_F = 6.91$}             
  & \multicolumn{3}{c}{$||\Delta W_q||_F = 3.57$}
  \\
  \bottomrule
  \end{tabular}
  \caption{The Frobenius norm of $U^\top W_qV^\top$ where $U$ and $V$ are the left/right top $r$ singular vector directions of either (1) $\Delta W_q$, (2) $W_q$, or (3) a random matrix. The weight matrices are taken from the 48th layer of GPT-3. 
  }
  \label{tab:w_delta_w_fro}
\end{table}

We draw \emph{several conclusions} from \autoref{tab:w_delta_w_fro}.
First, $\Delta W$ has a stronger correlation with $W$ compared to a random matrix, indicating that $\Delta W$ amplifies some features that are already in $W$.
Second, instead of repeating the top singular directions of $W$, $\Delta W$ only \emph{amplifies directions that are not emphasized in $W$}.
Third, the amplification factor is rather huge: $21.5\approx 6.91/0.32$ for $r=4$. 
See~\autoref{app:amplification_factor} for why $r=64$ has a smaller amplification factor.
We also provide a visualization in~\autoref{app:corr_w_delta_w} for how the correlation changes as we include more top singular directions from $W_q$.
This suggests that the low-rank adaptation matrix potentially \emph{amplifies the important features for specific downstream tasks that were learned but not emphasized in the general pre-training model}.

\section{Conclusion and Future Work}

Fine-tuning enormous language models is prohibitively expensive in terms of the hardware required and the storage/switching cost for hosting independent instances for different tasks.
We propose LoRA, an efficient adaptation strategy that neither introduces inference latency nor reduces input sequence length while retaining high model quality.
Importantly, it allows for quick task-switching when deployed as a service by sharing the vast majority of the model parameters.
While we focused on Transformer language models, the proposed principles are generally applicable to any neural networks with dense layers.

There are many directions for future works.
1) LoRA can be combined with other efficient adaptation methods, potentially providing orthogonal improvement.
2) The mechanism behind fine-tuning or LoRA is far from clear -- how are features learned during pre-training transformed to do well on downstream tasks?
We believe that LoRA makes it more tractable to answer this than full fine-tuning.
3) We mostly depend on heuristics to select the weight matrices to apply LoRA to.
Are there more principled ways to do it?
4) Finally, the rank-deficiency of $\Delta W$ suggests that $W$ could be rank-deficient as well, which can also be a source of inspiration for future works.

\bibliography{LORA, LoRA1}
\bibliographystyle{iclr2022_conference}

\appendix

\section{Large Language Models Still Need Parameter Updates}
\label{app:fewshot_vs_finetune}

Few-shot learning, or prompt engineering, is very advantageous when we only have a handful of training samples.
However, in practice, we can often afford to curate a few thousand or more training examples for performance-sensitive applications.
As shown in~\autoref{tab:gpt3_fs_ft_results}, fine-tuning improves the model performance drastically compared to few-shot learning on datasets large and small.
We take the GPT-3 few-shot result on RTE from the GPT-3 paper~\citep{brown_language_2020}.
For MNLI-matched, we use two demonstrations per class and six in-context examples in total.

\begin{table}[h]
\centering
\begin{tabular}{l|ccc}
\hline
\toprule
Method           & MNLI-m (Val. Acc./\%) & RTE (Val. Acc./\%) \\
\midrule
GPT-3 Few-Shot   & 40.6 & 69.0 \\
GPT-3 Fine-Tuned & 89.5 & 85.4 \\
\bottomrule
\end{tabular}
\caption{Fine-tuning significantly outperforms few-shot learning on GPT-3~\citep{brown_language_2020}.  }
\label{tab:gpt3_fs_ft_results}
\end{table}

\section{Inference Latency Introduced by Adapter Layers}
\label{app:adapter_latency}
Adapter layers are external modules added to a pre-trained model in a \textit{sequential} manner, whereas our proposal, LoRA, can be seen as external modules added in a parallel manner.
Consequently, adapter layers must be computed in addition to the base model, inevitably introducing additional latency.
While as pointed out in~\cite{ruckle2020adapterdrop}, the latency introduced by adapter layers can be mitigated when the model batch size and/or sequence length is large enough to full utilize the hardware parallelism.
We confirm their observation with a similar latency study on GPT-2 medium and point out that there are scenarios, notably online inference where the batch size is small, where the added latency can be significant.

\begin{figure}[h]
  \centering
    \includegraphics[width=0.9\textwidth]{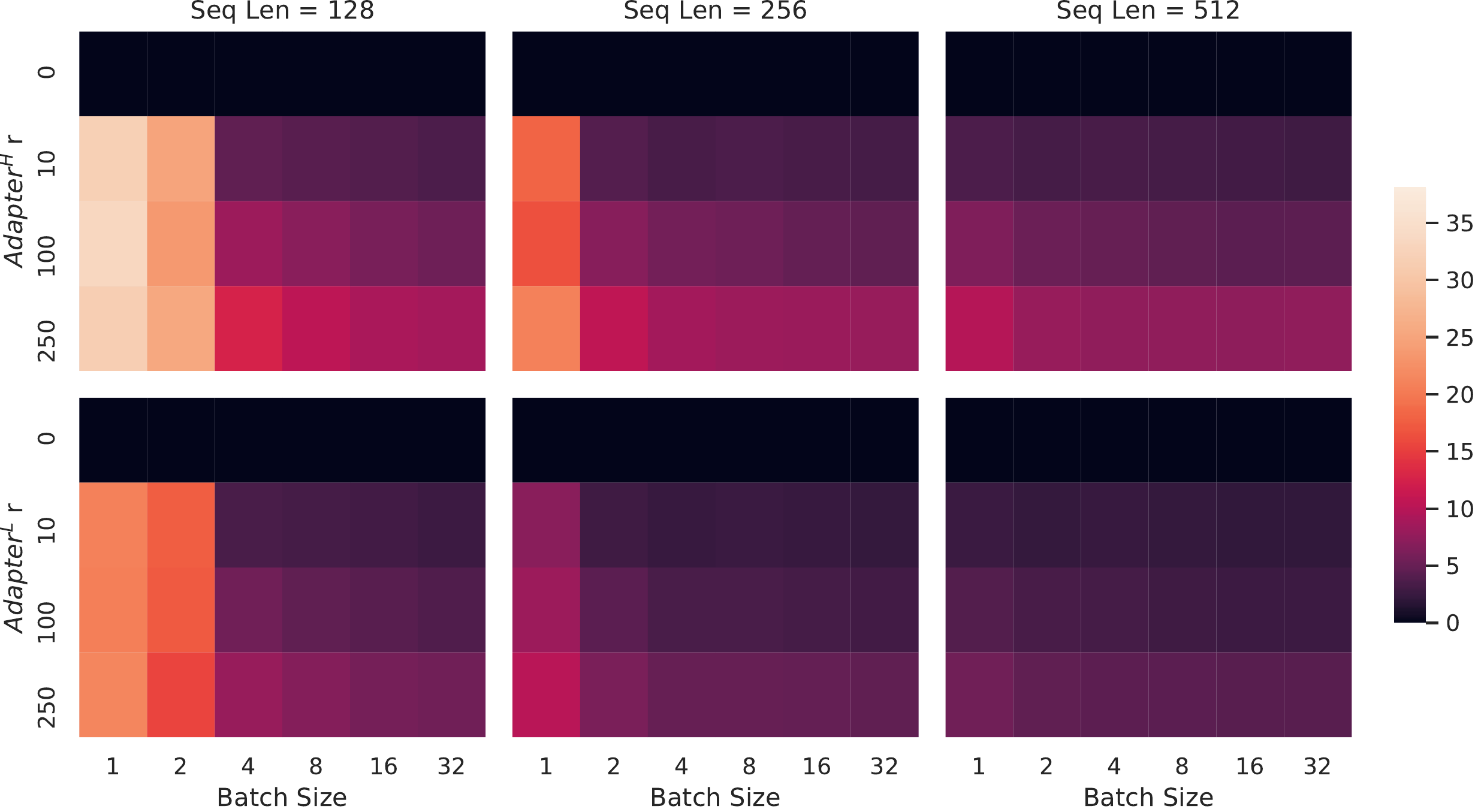}
    \caption{Percentage slow-down of inference latency compared to the no-adapter ($r=0$) baseline. The top row shows the result for $\text{Adapter}^{\text{H}}$ and the bottom row $\text{Adapter}^{\text{L}}$. Larger batch size and sequence length help to mitigate the latency, but the slow-down can be as high as over 30\% in an online, short-sequence-length scenario. We tweak the colormap for better visibility.}
    \label{fig:adapter_latency}
\end{figure}

We measure the latency of a single forward pass on an NVIDIA Quadro RTX8000 by averaging over 100 trials.
We vary the input batch size, sequence length, and the adapter bottleneck dimension $r$.
We test two adapter designs: the original one by~\cite{houlsby_parameter-efficient_2019}, which we call $\text{Adapter}^{\text{H}}$, and a recent, more efficient variant by~\cite{lin-etal-2020-exploring}, which we call $\text{Adapter}^{\text{L}}$.
See~\autoref{sec:expt_baselines} for more details on the designs.
We plot the slow-down in percentage compared to the no-adapter baseline in~\autoref{fig:adapter_latency}.

\section{Dataset Details}
\label{app:datasets}

\textbf{GLUE Benchmark}
is a wide-ranging collection of natural language understanding tasks.
It includes MNLI (inference,~\citet{williams-etal-2018-broad}), SST-2 (sentiment analysis,~\citet{socher-etal-2013-recursive}), MRPC (paraphrase detection,~\citet{dolan-brockett-2005-automatically}), CoLA (linguistic acceptability,~\citet{warstadt2018neural}), QNLI (inference,~\citet{DBLP:journals/corr/abs-1806-03822}), QQP\footnote{https://quoradata.quora.com/First-Quora-Dataset-Release-Question-Pairs} (question-answering), RTE (inference), and STS-B (textual similarity,~\citet{Cer_2017}).
The broad coverage makes GLUE benchmark a standard metric to evaluate NLU models such as RoBERTa and DeBERTa.
The individual datasets are released under different permissive licenses.

\textbf{WikiSQL}
is introduced in ~\cite{DBLP:journals/corr/abs-1709-00103} and contains $56,355$/$8,421$ training/validation examples.
The task is to generate SQL queries from natural language questions and table schemata.
We encode context as $x = \{ \text{table schema}, \text{query} \}$ and target as $y = \{ \text{SQL} \}$.
The dataset is release under the BSD 3-Clause License.

\textbf{SAMSum}
is introduced in ~\cite{DBLP:journals/corr/abs-1911-12237} and contains $14,732$/$819$ training/test examples.
It consists of staged chat conversations between two people and corresponding abstractive summaries written by linguists.
We encode context as "{\textbackslash}n" concatenated utterances followed by a "{\textbackslash}n{\textbackslash}n", and target as $y = \{ \text{summary} \}$. 
The dataset is released under the non-commercial licence: Creative Commons BY-NC-ND 4.0.

\textbf{E2E NLG Challenge} 
was first introduced in ~\cite{novikova2017e2e} as a dataset for training end-to-end, data-driven natural language generation systems and is commonly used for data-to-text evaluation. The E2E dataset consists of roughly $42,000$ training, $4,600$ validation, and $4,600$ test examples from the restaurant domain. Each source table used as input can have multiple references. Each sample input $(x, y)$ consists of a sequence of slot-value pairs, along with a corresponding natural language reference text.
The dataset is released under Creative Commons BY-NC-SA 4.0.

\textbf{DART} 
is an open-domain data-to-text dataset described in \cite{nan2020dart}.
DART inputs are structured as sequences of ENTITY | RELATION | ENTITY triples.
With $~82K$ examples in total, DART is a significantly larger and more complex data-to-text task compared to E2E.
The dataset is released under the MIT license.

\textbf{WebNLG} 
is another commonly used dataset for data-to-text evaluation~\citep{gardent2017webnlg}. With $~22K$ examples in total WebNLG comprises 14 distinct categories, nine of which are seen during training. Since five of the 14 total categories are not seen during training, but are represented in the test set, evaluation is typically broken out by ``seen'' categories (S),  ``unseen'' categories (U) and ``all'' (A). Each input example is represented by a sequence of SUBJECT | PROPERTY | OBJECT triples.
The dataset is released under Creative Commons BY-NC-SA 4.0.

\section{Hyperparameters Used in Experiments}

\subsection{RoBERTa}
\label{app:hps_roberta}

We train using AdamW with a linear learning rate decay schedule.
We sweep learning rate, number of training epochs, and batch size for LoRA.
Following~\citet{liu2019roberta}, we initialize the LoRA modules to our best MNLI checkpoint when adapting to MRPC, RTE, and STS-B, instead of the usual initialization; the pre-trained model stays frozen for all tasks.
We report the median over 5 random seeds; the result for each run is taken from the best epoch.
For a fair comparison with the setup in~\cite{houlsby_parameter-efficient_2019} and~\cite{pfeiffer2021adapterfusion}, we restrict the model sequence length to 128 and used a fixed batch size for all tasks.
Importantly, we start with the pre-trained RoBERTa large model when adapting to MRPC, RTE, and STS-B, instead of a model already adapted to MNLI.
The runs with this restricted setup are marked with $\dagger$.
See the hyperparameters used in our runs in~\autoref{tab:hyper_roberta}.

\begin{table}[h]
    \footnotesize
    \addtolength{\tabcolsep}{-1pt}
    \centering
    \begin{tabular}{ll|cccccccc}
        \hline
        \toprule
        Method  & Dataset     & MNLI & SST-2 & MRPC & CoLA & QNLI & QQP & RTE & STS-B \\
        \midrule
                              & Optimizer   & \multicolumn{8}{c}{AdamW} \\
                              & Warmup Ratio & \multicolumn{8}{c}{0.06} \\
                              & LR Schedule & \multicolumn{8}{c}{Linear} \\
        \midrule
        \multirow{5}{*}{\makecell{RoBERTa base \\ LoRA}} \ 
                              & Batch Size & 16 & 16 & 16 & 32 & 32 & 16 & 32 & 16 \\
                              & \# Epochs & 30 & 60 & 30 & 80 & 25 & 25 & 80 & 40 \\
                              & Learning Rate & 5E-04 & 5E-04 & 4E-04 & 4E-04 & 4E-04 & 5E-04 & 5E-04 & 4E-04 \\
                              & LoRA Config. & \multicolumn{8}{c}{$r_q=r_v=8$} \\
                              & LoRA $\alpha$ & \multicolumn{8}{c}{8} \\
                              & Max Seq. Len. & \multicolumn{8}{c}{512} \\
        \midrule
        \multirow{5}{*}{\makecell{RoBERTa large \\ LoRA}} \ 
                              & Batch Size & 4 & 4 & 4 & 4 & 4 & 4 & 8 & 8 \\
                              & \# Epochs & 10 & 10 & 20 & 20 & 10 & 20 & 20 & 30 \\
                              & Learning Rate & 3E-04 & 4E-04 & 3E-04 & 2E-04 & 2E-04 & 3E-04 & 4E-04 & 2E-04 \\
                              & LoRA Config. & \multicolumn{8}{c}{$r_q=r_v=8$} \\
                              & LoRA $\alpha$ & \multicolumn{8}{c}{16} \\
                              & Max Seq. Len. & 128 & 128 & 512 & 128 & 512 & 512 & 512 & 512 \\
        \midrule
        \multirow{5}{*}{\makecell{RoBERTa large \\ LoRA$\dagger$}} \ 
                              & Batch Size & \multicolumn{8}{c}{4} \\
                              & \# Epochs & 10 & 10 & 20 & 20 & 10 & 20 & 20 & 10 \\
                          & Learning Rate & 3E-04 & 4E-04 & 3E-04 & 2E-04 & 2E-04 & 3E-04 & 4E-04 & 2E-04 \\
                              & LoRA Config. & \multicolumn{8}{c}{$r_q=r_v=8$} \\
                              & LoRA $\alpha$ & \multicolumn{8}{c}{16} \\
                              & Max Seq. Len. & \multicolumn{8}{c}{128} \\
        \midrule
        \multirow{4}{*}{\makecell{RoBERTa large \\ $\text{Adpt}^\text{P}$ (3M)$\dagger$}} \ 
                              & Batch Size & \multicolumn{8}{c}{32} \\
                              & \# Epochs & 10 & 20 & 20 & 20 & 10 & 20 & 20 & 20 \\
                              & Learning Rate & 3E-05 & 3E-05 & 3E-04 & 3E-04 & 3E-04 & 3E-04 & 3E-04 & 3E-04 \\
                              & Bottleneck $r$ & \multicolumn{8}{c}{64} \\
                              & Max Seq. Len. & \multicolumn{8}{c}{128} \\
        \midrule
        \multirow{4}{*}{\makecell{RoBERTa large \\ $\text{Adpt}^\text{P}$ (0.8M)$\dagger$}} \ 
                              & Batch Size & \multicolumn{8}{c}{32} \\
                              & \# Epochs & 5 & 20 & 20 & 20 & 10 & 20 & 20 & 20 \\
                              & Learning Rate & 3E-04 & 3E-04 & 3E-04 & 3E-04 & 3E-04 & 3E-04 & 3E-04 & 3E-04 \\
                              & Bottleneck $r$ & \multicolumn{8}{c}{16} \\
                              & Max Seq. Len. & \multicolumn{8}{c}{128} \\
        \midrule
        \multirow{4}{*}{\makecell{RoBERTa large \\ $\text{Adpt}^\text{H}$ (6M)$\dagger$}} \
                              & Batch Size & \multicolumn{8}{c}{32} \\
                              & \# Epochs & 10 & 5 & 10 & 10 & 5 & 20 & 20 & 10 \\
                              & Learning Rate & 3E-05 & 3E-04 & 3E-04 & 3E-04 & 3E-04 & 3E-04 & 3E-04 & 3E-04 \\
                              & Bottleneck $r$ & \multicolumn{8}{c}{64} \\
                              & Max Seq. Len. & \multicolumn{8}{c}{128} \\
        \midrule
        \multirow{4}{*}{\makecell{RoBERTa large \\ $\text{Adpt}^\text{H}$ (0.8M)$\dagger$}} \
                              & Batch Size & \multicolumn{8}{c}{32} \\
                              & \# Epochs & 10 & 5 & 10 & 10 & 5 & 20 & 20 & 10 \\
                              & Learning Rate & 3E-04 & 3E-04 & 3E-04 & 3E-04 & 3E-04 & 3E-04 & 3E-04 & 3E-04 \\
                              & Bottleneck $r$ & \multicolumn{8}{c}{8} \\
                              & Max Seq. Len. & \multicolumn{8}{c}{128} \\
        \bottomrule
    \end{tabular}
    \caption{The hyperparameters we used for RoBERTa on the GLUE benchmark.}
    \label{tab:hyper_roberta}
\end{table}

\subsection{DeBERTa}
\label{app:hps_deberta}

We again train using AdamW with a linear learning rate decay schedule.
Following~\cite{he2021deberta}, we tune learning rate, dropout probability, warm-up steps, and batch size.
We use the same model sequence length used by~\citep{he2021deberta} to keep our comparison fair.
Following~\citet{he2021deberta}, we initialize the LoRA modules to our best MNLI checkpoint when adapting to MRPC, RTE, and STS-B, instead of the usual initialization; the pre-trained model stays frozen for all tasks.
We report the median over 5 random seeds; the result for each run is taken from the best epoch.
See the hyperparameters used in our runs in~\autoref{tab:hyper_deberta}.

\begin{table}[h]
    \footnotesize
    \addtolength{\tabcolsep}{-1pt}
    \centering
    \begin{tabular}{ll|cccccccc}
        \hline
        \toprule
        Method  & Dataset     & MNLI & SST-2 & MRPC & CoLA & QNLI & QQP & RTE & STS-B \\
        \midrule
                              & Optimizer   & \multicolumn{8}{c}{AdamW} \\
                              & Warmup Ratio & \multicolumn{8}{c}{0.1} \\
                              & LR Schedule & \multicolumn{8}{c}{Linear} \\
        \midrule
        \multirow{5}{*}{\makecell{DeBERTa XXL \\ LoRA}} \ 
                              & Batch Size & 8 & 8 & 32 & 4 & 6 & 8 & 4 & 4 \\
                              & \# Epochs & 5 & 16 & 30 & 10 & 8 & 11 & 11 & 10 \\
                              & Learning Rate & 1E-04 & 6E-05 & 2E-04 & 1E-04 & 1E-04 & 1E-04 & 2E-04 & 2E-04 \\
                              & Weight Decay & 0 & 0.01 & 0.01 & 0 & 0.01 & 0.01 & 0.01 & 0.1 \\
                              & CLS Dropout & 0.15 & 0 & 0 & 0.1 & 0.1 & 0.2 & 0.2 & 0.2 \\
                              & LoRA Config. & \multicolumn{8}{c}{$r_q=r_v=8$} \\
                              & LoRA $\alpha$ & \multicolumn{8}{c}{8} \\
                              & Max Seq. Len. & 256 & 128 & 128 & 64 & 512 & 320 & 320 & 128 \\
        \bottomrule
    \end{tabular}
    \caption{The hyperparameters for DeBERTa XXL on tasks included in the GLUE benchmark.}
    \label{tab:hyper_deberta}
\end{table}

\subsection{GPT-2}
\label{app:hps_gpt2}
We train all of our GPT-2 models using AdamW~\citep{loshchilov2017decoupled} with a linear learning rate schedule for 5 epochs.
We use the batch size, learning rate, and beam search beam size described in~\cite{li_prefix-tuning_2021}.
Accordingly, we also tune the above hyperparameters for LoRA.
We report the mean over 3 random seeds; the result for each run is taken from the best epoch.
The hyperparameters used for LoRA in GPT-2 are listed in~\autoref{tab:hyper_gpt2}.
For those used for other baselines, see~\cite{li_prefix-tuning_2021}.
\begin{table}[h]
\centering
\begin{tabular}{l|ccc}
\hline
\toprule
Dataset & E2E & WebNLG & DART \\
\midrule
&\multicolumn{3}{c}{Training} \\
\midrule
Optimizer & \multicolumn{3}{c}{AdamW} \\
Weight Decay & 0.01 & 0.01 & 0.0\\
Dropout Prob & 0.1 & 0.1 & 0.0\\
Batch Size & \multicolumn{3}{c}{8} \\
\# Epoch & \multicolumn{3}{c}{5} \\
Warmup Steps & \multicolumn{3}{c}{500} \\
Learning Rate Schedule & \multicolumn{3}{c}{Linear} \\
Label Smooth & 0.1 & 0.1 & 0.0 \\
Learning Rate & \multicolumn{3}{c}{0.0002} \\
Adaptation & \multicolumn{3}{c}{$r_q=r_v=4$} \\
LoRA $\alpha$ & \multicolumn{3}{c}{32} \\
\midrule
&\multicolumn{3}{c}{Inference} \\
\midrule
Beam Size & \multicolumn{3}{c}{10} \\
Length Penalty & 0.9 & 0.8 & 0.8 \\
no repeat ngram size & \multicolumn{3}{c}{4} \\
\bottomrule
\end{tabular}
\caption{The hyperparameters for GPT-2 LoRA on E2E, WebNLG and DART.}
\label{tab:hyper_gpt2}
\end{table}

\subsection{GPT-3}
\label{app:hps_gpt3}
For all GPT-3 experiments, we train using AdamW~\citep{loshchilov2017decoupled} for 2 epochs with a batch size of 128 samples and a weight decay factor of 0.1.
We use a sequence length of 384 for WikiSQL~\citep{DBLP:journals/corr/abs-1709-00103}, 768 for MNLI~\citep{williams-etal-2018-broad}, and 2048 for SAMSum~\citep{DBLP:journals/corr/abs-1911-12237}.
We tune learning rate for all method-dataset combinations.
See~\autoref{app:hps_gpt3} for more details on the hyperparameters used.
For prefix-embedding tuning, we find the optimal $l_p$ and $l_i$ to be 256 and 8, respectively, totalling $3.2M$ trainable parameters. 
We use $l_p=8$ and $l_i=8$ for prefix-layer tuning with $20.2M$ trainable parameters to obtain the overall best performance.
We present two parameter budgets for LoRA: 4.7M ($r_q=r_v=1$ or $r_{v}=2$) and 37.7M ($r_q=r_v=8$ or $r_q=r_k=r_v=r_o=2$).
We report the best validation performance from each run.
The training hyperparameters used in our GPT-3 experiments are listed in~\autoref{tab:hyper_gpt3}.
\begin{table}[h]
\centering
\begin{tabular}{l|cccccc}
\hline
\toprule
Hyperparameters & Fine-Tune & PreEmbed & PreLayer & BitFit & $\text{Adapter}^\text{H}$ & LoRA \\
\midrule
Optimizer & \multicolumn{6}{c}{AdamW} \\
Batch Size   & \multicolumn{6}{c}{128} \\
\# Epoch & \multicolumn{6}{c}{2} \\
Warmup Tokens & \multicolumn{6}{c}{250,000} \\
LR Schedule & \multicolumn{6}{c}{Linear} \\
\midrule
Learning Rate & 5.00E-06 & 5.00E-04 & 1.00E-04 & 1.6E-03 & 1.00E-04 & 2.00E-04 \\

\bottomrule
\end{tabular}
\caption{The training hyperparameters used for different GPT-3 adaption methods. We use the same hyperparameters for all datasets after tuning learning rate.}
\label{tab:hyper_gpt3}
\end{table}

\section{Combining LoRA with Prefix Tuning}
\label{app:lora_plus}
LoRA can be naturally combined with existing prefix-based approaches.
In this section, we evaluate two combinations of LoRA and variants of prefix-tuning on WikiSQL and MNLI.

\textbf{LoRA+PrefixEmbed (LoRA+PE)} combines LoRA with prefix-embedding tuning, where we insert $l_p+l_i$ special tokens whose embeddings are treated as trainable parameters.
For more on prefix-embedding tuning, see~\autoref{sec:expt_baselines}.

\textbf{LoRA+PrefixLayer (LoRA+PL)} combines LoRA with prefix-layer tuning.
We also insert $l_p+l_i$ special tokens; however, instead of letting the hidden representations of these tokens evolve naturally, we replace them after every Transformer block with an input agnostic vector.
Thus, both the embeddings and subsequent Transformer block activations are treated as trainable parameters.
For more on prefix-layer tuning, see~\autoref{sec:expt_baselines}.

In \autoref{tab:gpt3_ft_results_detailed}, we show the evaluation results of LoRA+PE and LoRA+PL on WikiSQL and MultiNLI.
First of all, LoRA+PE significantly outperforms both LoRA and prefix-embedding tuning on WikiSQL, which indicates that LoRA is somewhat orthogonal to prefix-embedding tuning.
On MultiNLI, the combination of LoRA+PE doesn't perform better than LoRA, possibly because LoRA on its own  already achieves performance comparable to the human baseline.
Secondly, we notice that LoRA+PL performs slightly worse than LoRA even with more trainable parameters.
We attribute this to the fact that prefix-layer tuning is very sensitive to the choice of learning rate and thus makes the optimization of LoRA weights more difficult in LoRA+PL.

\section{Additional Empirical Experiments}
\label{app:extra_expt}

\subsection{Additional Experiments on GPT-2}
\label{app:gpt2_extra}

We also repeat our experiment on DART~\citep{nan2020dart} and WebNLG~\citep{gardent2017webnlg} following the setup of~\cite{li_prefix-tuning_2021}.
The result is shown in~\autoref{tab:gpt2_ft_dart}.
Similar to our result on E2E NLG Challenge, reported in~\autoref{sec:empirical}, LoRA performs better than or at least on-par with prefix-based approaches given the same number of trainable parameters.

\begin{table}[h]
\centering
\begin{tabular}{lc|ccc}
\hline
\toprule
Method & \# Trainable &  \multicolumn{3}{c}{DART} \\ %
       & Parameters & BLEU$\uparrow$ & MET$\uparrow$ & TER$\downarrow$  \\
\midrule
 \multicolumn{5}{c}{GPT-2 Medium}\\
Fine-Tune & 354M & 46.2 & \textbf{0.39} & \textbf{0.46}  \\
$\text{Adapter}^{\text{L}}$ & 0.37M & 42.4 & 0.36 & 0.48 \\
$\text{Adapter}^{\text{L}}$ & 11M & 45.2 & 0.38 & \textbf{0.46} \\
$\text{FT}^{\text{Top2}}$ & 24M & 41.0 & 0.34 & 0.56 \\
PrefLayer & 0.35M & 46.4 & 0.38 & \textbf{0.46}   \\
LoRA  & 0.35M & \textbf{47.1}\textsubscript{$\pm$.2}  & \textbf{0.39} & \textbf{0.46} \\
\midrule
 \multicolumn{5}{c}{GPT-2 Large}\\
Fine-Tune & 774M & 47.0 & \textbf{0.39} & 0.46  \\
$\text{Adapter}^{\text{L}}$ & 0.88M & 45.7\textsubscript{$\pm$.1} & 0.38 & 0.46 \\
$\text{Adapter}^{\text{L}}$ & 23M & 47.1\textsubscript{$\pm$.1} & \textbf{0.39} & \textbf{0.45} \\
PrefLayer & 0.77M & 46.7 & 0.38 & \textbf{0.45} \\
LoRA & 0.77M & \textbf{47.5}\textsubscript{$\pm$.1} & \textbf{0.39} &  \textbf{0.45} \\
\bottomrule
\end{tabular}
\caption{GPT-2 with different adaptation methods on DART. The variances of MET and TER are less than $0.01$ for all adaption approaches. }
\label{tab:gpt2_ft_dart}
\end{table}

\begin{table}[h]
\centering
\begin{tabular}{l|ccc|ccc|ccc}
\hline
\toprule
Method & \multicolumn{9}{c}{WebNLG} \\ %
       & \multicolumn{3}{c}{BLEU$\uparrow$}  &  \multicolumn{3}{c}{MET$\uparrow$} & \multicolumn{3}{c}{TER$\downarrow$ } \\
       & U & S & A & U & S & A & U & S & A \\
\midrule
&  \multicolumn{9}{c}{GPT-2 Medium}\\
Fine-Tune (354M)                    & 27.7                                  & \textbf{64.2}                         & 46.5                                  & .30           & \textbf{.45}  & .38           &.76            & \textbf{.33}           & .53  \\
$\text{Adapter}^{\text{L}}$ (0.37M)  & 45.1                                  & 54.5                                  & 50.2                                  & .36           & .39           & .38           & .46           & .40           & .43 \\
$\text{Adapter}^{\text{L}}$ (11M)   & \textbf{48.3}                         & 60.4                                  & 54.9                                  & \textbf{.38}  & .43           & \textbf{.41}  & \textbf{.45}  & .35           &  \textbf{.39} \\
$\text{FT}^{\text{Top2}}$ (24M)     & 18.9                                  & 53.6                                  & 36.0                                  & .23           & .38           & .31           & .99          & .49           & .72 \\
Prefix (0.35M)                      & 45.6                                  & 62.9                                  & 55.1                                  & \textbf{.38}  & .44           & \textbf{.41}  & .49           & .35           & .40   \\
LoRA (0.35M)                        & 46.7\textsubscript{$\pm$.4}           & 62.1\textsubscript{$\pm$.2}           & \textbf{55.3}\textsubscript{$\pm$.2}  & \textbf{.38}  & .44           & \textbf{.41}  & .46           & \textbf{.33}  & \textbf{.39} \\
\midrule
& \multicolumn{9}{c}{GPT-2 Large}\\
Fine-Tune (774M)                    & 43.1                                  & 65.3                                  & 55.5                                  & .38           & \textbf{.46}  & .42           & .53           & .33           & .42  \\
$\text{Adapter}^{\text{L}}$ (0.88M)  & \textbf{49.8}\textsubscript{$\pm$.0}  & 61.1\textsubscript{$\pm$.0}           & 56.0\textsubscript{$\pm$.0}           & .38           & .43           & .41           & \textbf{.44}  & .35           &  .39 \\
$\text{Adapter}^{\text{L}}$ (23M)   & 49.2\textsubscript{$\pm$.1}           & 64.7\textsubscript{$\pm$.2}           & \textbf{57.7}\textsubscript{$\pm$.1}  & \textbf{.39}  & \textbf{.46}  & \textbf{.43}  & .46           & .33           &  .39 \\
Prefix (0.77M)                      & 47.7                                  & 63.4                                  & 56.3                                  & \textbf{.39}  & .45           & .42           & .48           & .34           & .40 \\
LoRA (0.77M)                        & 48.4\textsubscript{$\pm$.3}           & 64.0\textsubscript{$\pm$.3}           & 57.0\textsubscript{$\pm$.1}           & \textbf{.39}  & .45           & .42           & .45           & \textbf{.32}  & \textbf{.38} \\
\bottomrule
\end{tabular}
\caption{GPT-2 with different adaptation methods on WebNLG. The variances of MET and TER are less than $0.01$ for all the experiments we ran. ``U'' indicates unseen categories, ``S'' indicates seen categories, and ``A'' indicates all categories in the test set of WebNLG.}
\label{tab:gpt2_ft_webnlg}
\end{table}

\subsection{Additional Experiments on GPT-3}
\label{app:gpt3_extra}
We present additional runs on GPT-3 with different adaptation methods in~\autoref{tab:gpt3_ft_results_detailed}.
The focus is on identifying the trade-off between performance and the number of trainable parameters.
\begin{table}[h]
\centering
\begin{tabular}{l|c|c|c|c}
\hline
\toprule
Method    & Hyperparameters & \# Trainable Parameters &  WikiSQL & MNLI-m \\
\midrule
Fine-Tune & - & 175B &  73.8 & 89.5 \\
\midrule
\multirow{5}{*}{PrefixEmbed} & $l_p=32, l_i=8$ & 0.4 M & 55.9 & 84.9 \\
                             & $l_p=64, l_i=8$ & 0.9 M & 58.7 & 88.1 \\
                             & $l_p=128, l_i=8$ & 1.7 M & 60.6 & 88.0 \\
                             & $l_p=256, l_i=8$ & 3.2 M & 63.1 & 88.6 \\
                             & $l_p=512, l_i=8$ & 6.4 M & 55.9 & 85.8 \\
\midrule
\multirow{5}{*}{PrefixLayer} 
                             & $l_p=2, l_i=2$ & 5.1 M & 68.5 & 89.2 \\
                             & $l_p=8, l_i=0$ & 10.1 M & 69.8 & 88.2 \\
                             & $l_p=8, l_i=8$ & 20.2 M & 70.1 & 89.5 \\
                             & $l_p=32, l_i=4$ & 44.1 M & 66.4 & 89.6 \\
                             & $l_p=64, l_i=0$ & 76.1 M & 64.9 & 87.9 \\
\midrule
\multirow{5}{*}{$\text{Adapter}^{\text{H}}$} 
                             & $r=1$ & 7.1 M & 71.9 & 89.8 \\
                             & $r=4$ & 21.2 M & 73.2 & 91.0 \\
                             & $r=8$ & 40.1 M & 73.2 & 91.5 \\
                             & $r=16$& 77.9 M & 73.2 & 91.5 \\
                             & $r=64$& 304.4 M & 72.6 & 91.5 \\
\midrule
\multirow{5}{*}{LoRA}        & $r_{v}=2 $ & 4.7 M & 73.4 & \textbf{91.7} \\
                             & $r_q=r_v=1 $ & 4.7 M & 73.4 & 91.3 \\
                             & $r_q=r_v=2 $ & 9.4 M & 73.3 & 91.4 \\
                             & $r_q=r_k=r_v=r_o=1$ & 9.4 M & 74.1 & 91.2 \\
                             & $r_q=r_v=4 $ & 18.8 M & 73.7 & 91.3 \\
                             & $r_q=r_k=r_v=r_o=2$ & 18.8 M & 73.7 & \textbf{91.7} \\
                             & $r_q=r_v=8$ & 37.7 M & 73.8 & \textbf{91.6} \\
                             & $r_q=r_k=r_v=r_o=4$ & 37.7 M & 74.0 & \textbf{91.7} \\
                             & $r_q=r_v=64$ & 301.9 M & 73.6 & 91.4 \\
                             & $r_q=r_k=r_v=r_o=64$ & 603.8 M & 73.9 & 91.4 \\
\midrule
\multirow{3}{*}{LoRA+PE}     & $r_q=r_v=8, l_p=8, l_i=4$ & 37.8 M & 75.0 & 91.4\\
                             & $r_q=r_v=32, l_p=8, l_i=4$ &151.1 M & \textbf{75.9} & 91.1\\
                             & $r_q=r_v=64, l_p=8, l_i=4$ &302.1 M & \textbf{76.2}  & 91.3\\
\midrule
\multirow{1}{*}{LoRA+PL} & $r_q=r_v=8, l_p=8, l_i=4$ & 52.8 M & 72.9 & 90.2\\
\bottomrule
\end{tabular}
\caption{Hyperparameter analysis of different adaptation approaches on WikiSQL and MNLI. Both prefix-embedding tuning (PrefixEmbed) and prefix-layer tuning (PrefixLayer) perform worse as we increase the number of trainable parameters, while LoRA's performance stabilizes. Performance is measured in validation accuracy.}
\label{tab:gpt3_ft_results_detailed}
\end{table}

\subsection{Low-Data Regime}
\label{app:low_data}
To evaluate the performance of different adaptation approaches in the low-data regime. we randomly sample 100, 1k and 10k training examples from the full training set of MNLI to form the low-data MNLI-$n$ tasks.
In \autoref{tab:gpt3-few-shot-mnli}, we show the performance of different adaptation approaches on MNLI-$n$. 
To our surprise, PrefixEmbed and PrefixLayer performs very poorly on MNLI-100 dataset, with PrefixEmbed performing only slightly better than random chance (37.6\% vs. 33.3\%).
PrefixLayer performs better than PrefixEmbed but is still significantly worse than Fine-Tune or LoRA on MNLI-100.
The gap between prefix-based approaches and LoRA/Fine-tuning becomes smaller as we increase the number of training examples, which might suggest that prefix-based approaches are not suitable for low-data tasks in GPT-3.
LoRA achieves better performance than fine-tuning on both MNLI-100 and MNLI-Full, and comparable results on MNLI-1k and MNLI-10K considering the ($\pm0.3$) variance due to random seeds. 
\begin{table}[h]
\centering
\begin{tabular}{l|cccc}
\hline
\toprule
Method & MNLI(m)-100 &  MNLI(m)-1k &  MNLI(m)-10k &  MNLI(m)-392K \\
\midrule
GPT-3 (Fine-Tune)  & 60.2  & \textbf{85.8} & 88.9 & 89.5 \\
GPT-3 (PrefixEmbed) & 37.6  & 75.2  & 79.5 & 88.6 \\
GPT-3 (PrefixLayer) & 48.3  & 82.5  & 85.9 & 89.6 \\
GPT-3 (LoRA)         & \textbf{63.8}   & 85.6  & \textbf{89.2} & \textbf{91.7} \\
\bottomrule
\end{tabular}
\caption{Validation accuracy of different methods on subsets of MNLI using GPT-3 175B. MNLI-$n$ describes a subset with $n$ training examples. We evaluate with the full validation set. LoRA performs exhibits favorable sample-efficiency compared to other methods, including fine-tuning.}
\label{tab:gpt3-few-shot-mnli}
\end{table}

The training hyperparameters of different adaptation approaches on MNLI-{n} are reported in \autoref{tab:hyper_gpt3_low}.
We use a smaller learning rate for PrefixLayer on the MNLI-100 set, as the training loss does not decrease with a larger learning rate.

\begin{table}[h]
\centering
\begin{tabular}{l|c|cccc}
\hline
\toprule
Hyperparameters & Adaptation & MNLI-100 & MNLI-1k & MNLI-10K & MNLI-392K \\
\midrule
Optimizer & - & \multicolumn{4}{c}{AdamW} \\
Warmup Tokens & - & \multicolumn{4}{c}{250,000} \\
LR Schedule & - & \multicolumn{4}{c}{Linear} \\
Batch Size & - & 20 & 20 & 100 & 128 \\
\# Epoch & - & 40 & 40 & 4 & 2 \\
\midrule
\multirow{4}{*}{Learning Rate} & FineTune & \multicolumn{4}{c}{5.00E-6} \\
                               & PrefixEmbed & 2.00E-04 & 2.00E-04 & 4.00E-04 & 5.00E-04 \\
                               & PrefixLayer & 5.00E-05 & 5.00E-05 & 5.00E-05 & 1.00E-04 \\
                               & LoRA & \multicolumn{4}{c}{2.00E-4} \\
\midrule
                            & PrefixEmbed $l_p$ & 16 & 32 & 64 & 256 \\
 Adaptation-                & PrefixEmbed $l_i$ & \multicolumn{4}{c}{8} \\
 Specific                   & PrefixTune & \multicolumn{4}{c}{$l_p=l_i=8$} \\
                            & LoRA & \multicolumn{4}{c}{$r_q=r_v=8$} \\
\bottomrule
\end{tabular}
\caption{The hyperparameters used for different GPT-3 adaptation methods on MNLI(m)-$n$.}
\label{tab:hyper_gpt3_low}
\end{table}

\section{Measuring Similarity Between Subspaces}
\label{app:grassmann_distance}

In this paper we use the measure $\phi(A, B, i, j) = \psi(U_A^i, U_B^j) = \frac{\| U_A^{i\top } U_B \|_F^2}{\min \{i, j\}}$ to measure the subspace similarity between two column orthonormal matrices $U_A^i \in \mathbb{R}^{d \times i}$ and $U_B^j \in \mathbb{R}^{d \times j}$, obtained by taking columns of the left singular matrices of $A$ and $B$.
We point out that this similarity is simply a reverse of the standard Projection Metric that measures distance between subspaces~\cite{dist}.

To be concrete, let the singular values of $ U_A^{i\top } U_B^j$ to be $\sigma_1, \sigma_2, \cdots, \sigma_{p}$ where $p = \min \{i, j \}$. We know that the Projection Metric~\cite{dist} is defined as:
$$d(U_A^i, U_B^j) = \sqrt{ p - \sum_{i = 1}^p \sigma_i^2} \in [0, \sqrt{p}]$$

where our similarity is defined as:
$$\phi(A, B, i, j) = \psi(U_A^i, U_B^j) = \frac{\sum_{i = 1}^p \sigma_i^2}{p}= \frac{1}{p} \left( 1 - d(U_A^i, U_B^j)^2 \right)$$

This similarity satisfies that if $U_A^i$ and $U_B^j$ share the same column span, then $\phi(A, B, i, j) = 1$.
If they are completely orthogonal, then $\phi(A, B, i, j) = 0$. Otherwise, $\phi(A, B, i, j) \in (0, 1)$.

\section{Additional Experiments on Low-Rank Matrices}
We present additional results from our investigation into the low-rank update matrices.
\subsection{Correlation between LoRA Modules}
\label{app:corr_lora}
\begin{figure}[h]
  \centering
    \includegraphics[width=0.99\textwidth]{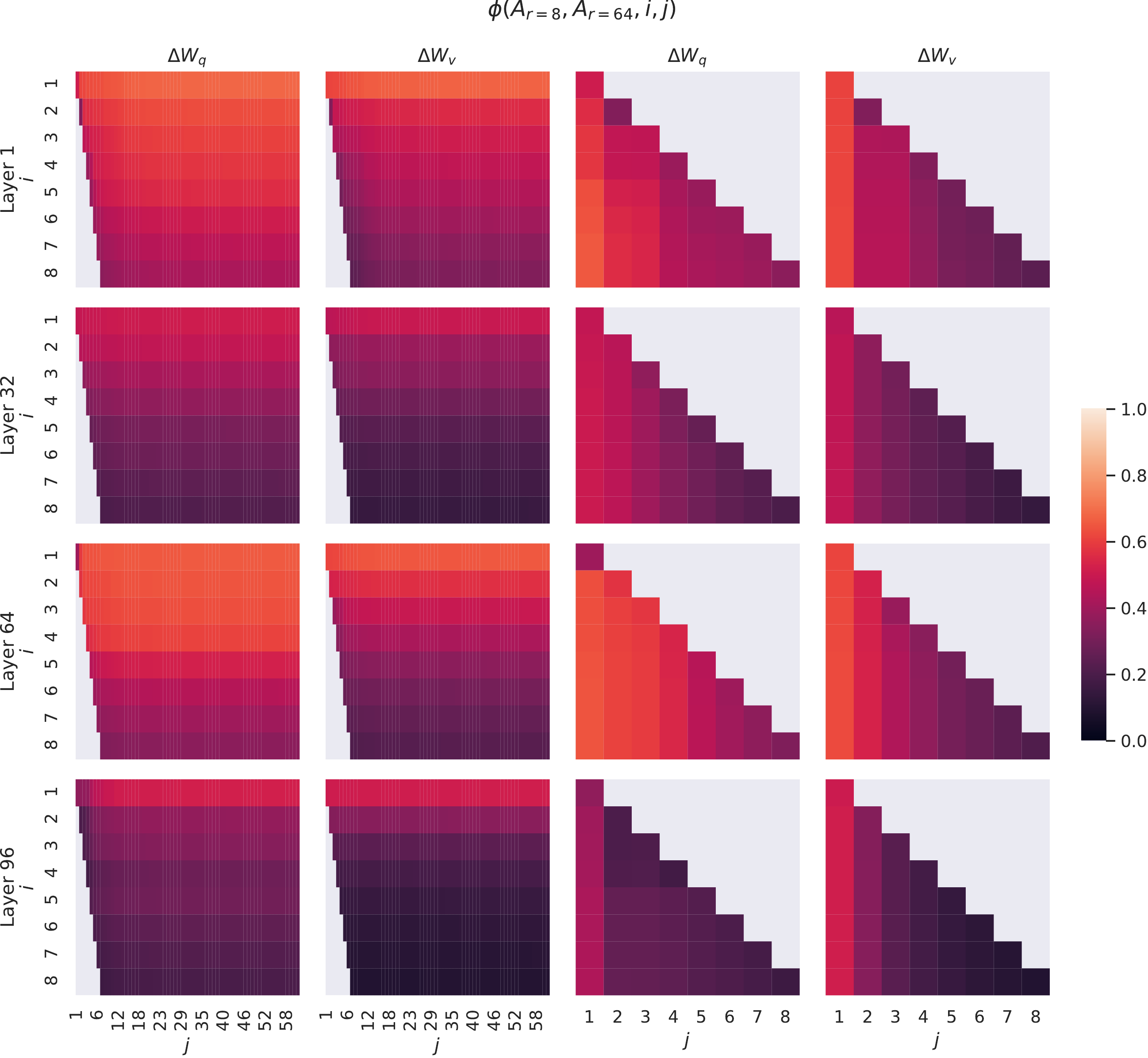}
    \caption{Normalized subspace similarity between the column vectors of $A_{r=8}$ and $A_{r=64}$ for both $\Delta W_q$ and $\Delta W_v$ from the 1st, 32nd, 64th, and 96th layers in a 96-layer Transformer.}
    \label{fig:qv8_qv64_more_layers}
\end{figure}

\begin{figure}[h]
  \centering
    \includegraphics[width=0.99\textwidth]{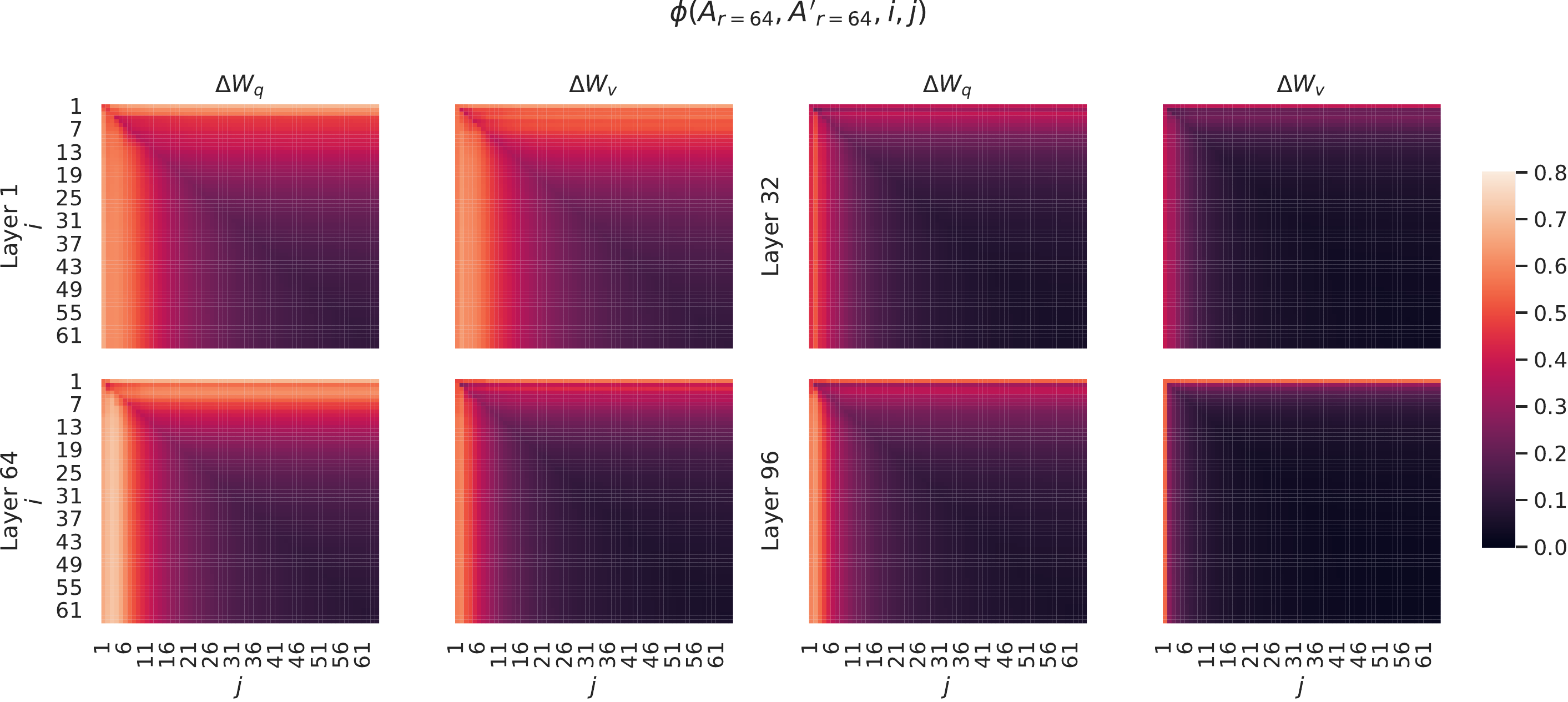}
    \caption{Normalized subspace similarity between the column vectors of $A_{r=64}$ from two randomly seeded runs, for both $\Delta W_q$ and $\Delta W_v$ from the 1st, 32nd, 64th, and 96th layers in a 96-layer Transformer.}
    \label{fig:UA_across_random_seeds_more_layers}
\end{figure}

See~\autoref{fig:qv8_qv64_more_layers} and \autoref{fig:UA_across_random_seeds_more_layers} for how the results presented in~\autoref{fig:qv8_qv64} and \autoref{fig:UA_across_random_seeds} generalize to other layers.

\subsection{Effect of $r$ on GPT-2}
\label{app:gpt2_effect_r}

We repeat our experiment on the effect of $r$ (\autoref{sec:effect_of_r}) in GPT-2.
Using the E2E NLG Challenge dataset as an example, we report the validation loss and test metrics achieved by different choices of $r$ after training for 26,000 steps.
We present our result in~\autoref{tab:effect_of_r_gpt2}.
The optimal rank for GPT-2 Medium is between 4 and 16 depending on the metric used, which is similar to that for GPT-3 175B.
Note that the relationship between model size and the optimal rank for adaptation is still an open question.

\begin{table}[h]
  \centering
  \begin{tabular}{c|llllll}
  \toprule
Rank $r$& val\_loss     & BLEU            & NIST            & METEOR          & ROUGE\_L        & CIDEr           \\
\midrule
1    & 1.23          & 68.72          & 8.7215          & 0.4565          & 0.7052          & 2.4329          \\
2    & 1.21          & 69.17          & 8.7413          & 0.4590          & 0.7052          & 2.4639          \\
4    & 1.18          & \textbf{70.38} & \textbf{8.8439} & \textbf{0.4689} & 0.7186          & \textbf{2.5349} \\
8    & 1.17          & 69.57          & 8.7457          & 0.4636          & \textbf{0.7196} & 2.5196          \\
16   & \textbf{1.16} & 69.61          & 8.7483          & 0.4629          & 0.7177          & 2.4985          \\
32   & \textbf{1.16} & 69.33          & 8.7736          & 0.4642          & 0.7105          & 2.5255          \\
64   & \textbf{1.16} & 69.24          & 8.7174          & 0.4651          & 0.7180          & 2.5070          \\
128  & \textbf{1.16} & 68.73          & 8.6718          & 0.4628          & 0.7127          & 2.5030          \\
256  & \textbf{1.16} & 68.92          & 8.6982          & 0.4629          & 0.7128          & 2.5012          \\
512  & \textbf{1.16} & 68.78          & 8.6857          & 0.4637          & 0.7128          & 2.5025          \\
1024 & 1.17          & 69.37          & 8.7495          & 0.4659          & 0.7149          & 2.5090          \\
  \bottomrule
  \end{tabular}
  \caption{Validation loss and test set metrics on E2E NLG Challenge achieved by LoRA with different rank $r$ using GPT-2 Medium. Unlike on GPT-3 where $r=1$ suffices for many tasks, here the performance peaks at $r=16$ for validation loss and $r=4$ for BLEU, suggesting the GPT-2 Medium has a similar intrinsic rank for adaptation compared to GPT-3 175B. Note that some of our hyperparameters are tuned on $r=4$, which matches the parameter count of another baseline, and thus might not be optimal for other choices of $r$.}
  \label{tab:effect_of_r_gpt2}
\end{table}

\subsection{Correlation between $W$ and $\Delta W$}
\label{app:corr_w_delta_w}

\begin{figure}[h]
  \centering
    \includegraphics[width=0.99\textwidth]{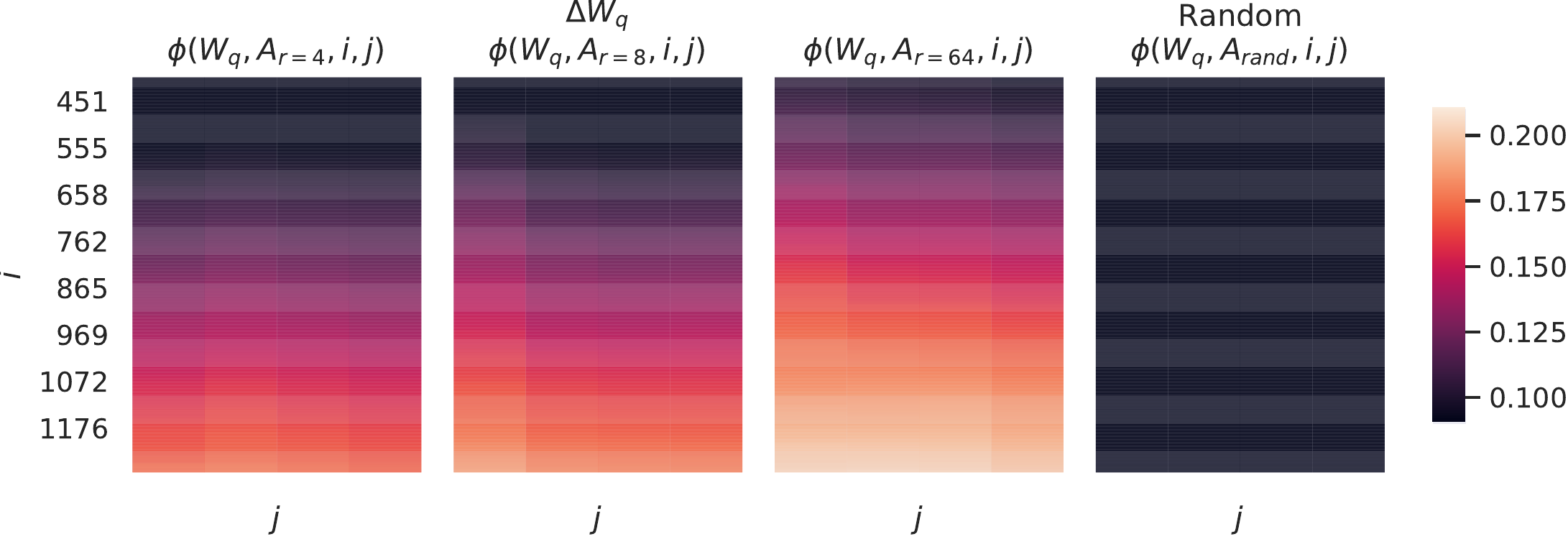}
    \caption{Normalized subspace similarity between the singular directions of $W_q$ and those of $\Delta W_q$ with varying $r$ and a random baseline. $\Delta W_q$ amplifies directions that are important but not emphasized in $W$. $\Delta W$ with a larger $r$ tends to pick up more directions that are already emphasized in $W$.}
    \label{fig:w_vs_delta_w}
\end{figure}

See~\autoref{fig:w_vs_delta_w} for the normalized subspace similarity between $W$ and $\Delta W$ with varying $r$.

Note again that $\Delta W$ does not contain the top singular directions of $W$, since the similarity between the top 4 directions in $\Delta W$ and the top-10\% of those in $W$ barely exceeds 0.2. This gives evidence that $\Delta W$ contains those ``task-specific'' directions that are otherwise \emph{not} emphasized in $W$.

An interesting next question to answer, is how ``strong'' do we need to amplify those task-specific directions, in order for the model adaptation to work well?

\subsection{Amplification Factor}
\label{app:amplification_factor}

One can naturally consider a \emph{feature amplification factor} as the \emph{ratio} $\frac{\|\Delta W\|_F}{\|U^\top WV^\top\|_F}$, where $U$ and $V$ are the left- and right-singular matrices of the SVD decomposition of $\Delta W$. (Recall $UU^\top WV^\top V$ gives the ``projection'' of $W$ onto the subspace spanned by $\Delta W$.)

Intuitively, when $\Delta W$ mostly contains task-specific directions, this quantity measures how much of them are amplified by $\Delta W$.
As shown in~\autoref{sec:compare_delta_w_to_w}, for $r=4$, this amplification factor is as large as 20. In other words, there are (generally speaking) four feature directions in each layer (out of the entire feature space from the pre-trained model $W$), that need to be amplified by a very large factor 20, in order to achieve our reported accuracy for the downstream specific task. And, one should expect a very different set of feature directions  to be amplified for each different downstream task.

One may notice, however, for $r=64$, this amplification factor is only around 2, meaning that \emph{most} directions learned in $\Delta W$ with $r=64$ are \emph{not} being amplified by much.
This should not be surprising, and in fact gives evidence (once again) that the intrinsic rank \emph{needed} to represent the ``task-specific directions'' (thus for model adaptation) is low.
In contrast, those directions in the rank-4 version of $\Delta W$ (corresponding to $r=4$) are amplified by a much larger factor 20.

\end{document}